\documentclass{article}

\usepackage{microtype}
\usepackage{graphicx}
\usepackage{subfigure}
\usepackage{booktabs} 

\usepackage{hyperref}

\usepackage[accepted]{mlsys2026}

\mlsystitlerunning{Flashlight: A PyTorch Compiler Framework for Accelerating Attention Variants}

\AddToShipoutPictureBG*{
  \AtPageUpperLeft{
    \hspace*{\paperwidth}
    \raisebox{-68pt}{
      \llap{
        \href{https://www.acm.org/publications/policies/artifact-review-and-badging-current}{
          \includegraphics[height=65pt]{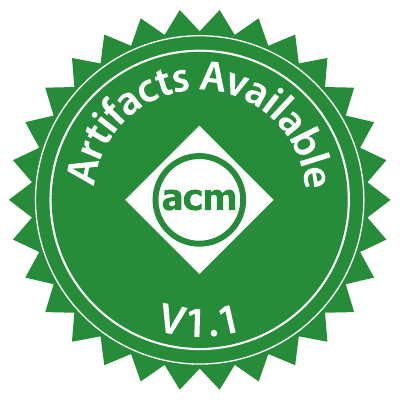}}
        \hspace{1pt}
        \href{https://www.acm.org/publications/policies/artifact-review-and-badging-current}{
          \includegraphics[height=65pt]{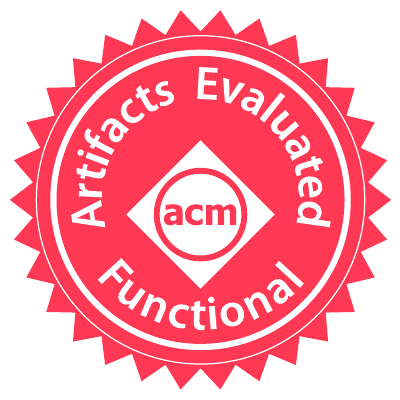}}
        \hspace{1pt}
        \href{https://www.acm.org/publications/policies/artifact-review-and-badging-current}{
          \includegraphics[height=65pt]{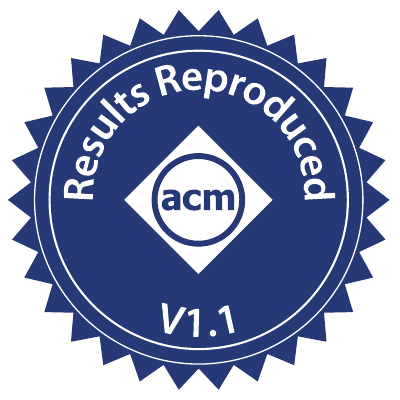}}
        \hspace{80pt}
      }
    }
  }
}

\usepackage{amsmath,amssymb,amsfonts}
\usepackage{mdframed}
\usepackage{graphicx}  
\usepackage{textcomp}
\usepackage{xcolor}
\def\BibTeX{{\rm B\kern-.05em{\sc i\kern-.025em b}\kern-.08em
    T\kern-.1667em\lower.7ex\hbox{E}\kern-.125emX}}

\usepackage[utf8]{inputenc} 
\usepackage[T1]{fontenc}    
\usepackage{hyperref}       
\usepackage{url}            
\usepackage{booktabs}       
\usepackage{amsfonts}       
\usepackage{nicefrac}       
\usepackage{microtype}      
\usepackage{xcolor}         
\usepackage{listings}

\usepackage[english]{babel} 
\usepackage[all]{nowidow} 

\usepackage{enumitem}

\usepackage{tikz}
\usetikzlibrary{positioning, fit, calc, shapes, arrows, arrows.meta}
\tikzset{font=\small, node distance=5pt, >=latex}

\usepackage{minted}  

\usepackage[colorinlistoftodos,
    textcolor=orange,
    backgroundcolor=white,
    bordercolor=orange]{todonotes}

\usepackage{arydshln}
\usepackage[normalem]{ulem}
\usepackage{comment}

\newcommand{\flashlight}[0]{\textsc{FlashLight}}

\newcommand\vVecLen{N}
\newcommand\vVec{\mathbf{x}}
\newcommand\vSoftmax{\mathrm{softmax}}

\newtheorem{definition}{Definition}

\newenvironment{closeitemize}
{\begin{itemize}
    \setlength{\itemsep}{1pt}
    \setlength{\parskip}{0pt}
    \setlength{\parsep}{0pt}}
{\end{itemize}}

\newenvironment{closeenumerate}
{\begin{enumerate}
    \setlength{\itemsep}{1pt}
    \setlength{\parskip}{0pt}
    \setlength{\parsep}{0pt}}
{\end{enumerate}}

\newcommand{\revised}[1]{#1}

\newcommand{\withcolon}[1]{#1:}

\begin{document}

\twocolumn[
\mlsystitle{\textsc{Flashlight}: PyTorch Compiler Extensions to Accelerate Attention Variants}




\begin{mlsysauthorlist}
\mlsysauthor{Bozhi You}{ut,intern,google}
\mlsysauthor{Irene Wang}{gatech,intern}
\mlsysauthor{Zelal Su Mustafaoglu}{ut}
\mlsysauthor{Abhinav Jangda}{msr}
\mlsysauthor{Angélica Moreira}{msr}
\mlsysauthor{Roshan Dathathri}{msr}
\mlsysauthor{Divya Mahajan}{gatech}
\mlsysauthor{Keshav Pingali}{ut}
\end{mlsysauthorlist}

\mlsysaffiliation{ut}{Department of Computer Science, University of Texas at Austin, Austin, USA}
\mlsysaffiliation{gatech}{School of Computer Science, Georgia Institute of Technology, Atlanta, USA}
\mlsysaffiliation{msr}{Microsoft Research, Redmond, USA}
\mlsysaffiliation{intern}{Work started while at Microsoft Research}
\mlsysaffiliation{google}{Now at Google}

\mlsyscorrespondingauthor{Bozhi You}{youbozhi@cs.utexas.edu}
\mlsyscorrespondingauthor{Roshan Dathathri}{roshan.dathathri@microsoft.com}

\mlsyskeywords{Machine Learning, MLSys}

\vskip 0.3in

\begin{abstract}
Attention is a fundamental building block of large language models (LLMs), so there have been many efforts to implement it efficiently. For example, FlashAttention leverages tiling and kernel fusion to optimize attention. Recently, a number of variants of attention have been introduced to enhance model quality or efficiency. Supporting them efficiently remains difficult since they usually require specialized kernels or hand-tuned implementations. FlexAttention recently addressed part of this gap by using static programming templates to support FlashAttention-like kernels for a subset of attention variants.

In this paper, we introduce \flashlight{}, a compiler-native framework within the PyTorch ecosystem that automatically generates fused, FlashAttention-style kernels for arbitrary attention-based programs, without relying on static templates or predefined kernel specializations. \flashlight{} leverages PyTorch’s compilation workflow to fuse and tile attention computations transparently, enabling efficient execution for diverse attention patterns. Not only does it support all variants expressible in the FlexAttention model but it also handles more general, data-dependent attention formulations that are beyond the capabilities of FlexAttention.
Our results show that \flashlight{} produces kernels with competitive or superior performance to FlexAttention, while offering the flexibility of native PyTorch code, enabling developers to rapidly explore  new attention models without sacrificing performance. 
\revised{\flashlight{} is open source and available as a fork of PyTorch at \url{https://github.com/bozhiyou/pytorch-flashlight}.}
\end{abstract}
]



\printAffiliationsAndNotice{}  

\section{Introduction}\label{sec:introduction}

Optimizing attention is crucial for accelerating training and inference in machine learning pipelines. For instance, FlashAttention~\cite{dao2022flashattentionfastmemoryefficientexact,dao2023flashattention2fasterattentionbetter,shah2024flashattention3fastaccurateattention} improves performance by tiling and fusing multiple operations into a single kernel, thereby reducing memory reads/writes and launch overhead to enhance data locality and GPU utilization. 
Variants of attention such as differential attention~\cite{ye2024differentialtransformer}, row/column-wise gated self-attention in AlphaFold's~\cite{alpafold} Evoformer, AlphaFold's Invariant Point Attention (IPA), and Rectified Sparse Attention (RSA)~\cite{rectifiedsparseattention} have been introduced to improve model quality and reduce hallucinations.
\revised{Handcrafted kernels like FlashAttention cannot be used for new attention variants}.
Therefore, achieving high performance for new attention models often requires significant engineering effort, leaving many variants constrained by slower PyTorch implementations that lack fused, memory-efficient execution, and hindering their adoption.

FlexAttention~\cite{dong2024flexattentionprogrammingmodel,2024flexattnblog} 
\revised{and FlashInfer~\cite{ye2025flashinferefficientcustomizableattention} attempt} to circumvent this problem by providing a static, higher-order template that captures a range of known attention variants. 
\revised{FlexAttention builds on PyTorch’s compiler infrastructure, using the TorchInductor to generate a fused Triton~\cite{Tillet2019Triton} kernel from template-specialized Triton kernels. 
In contrast, FlashInfer uses just-in-time compilation to generate a fused CUDA kernel from template-specialized CUDA kernels. 
Nevertheless, both FlexAttention and FlashInfer have limited expressiveness since they require} programmers to express their attention variant in terms of their template. In particular, variants such as differential attention, row/column-wise gated self-attention, IPA, and RSA do not fit their template, preventing them from achieving the performance of handcrafted implementations.

Compiler-driven optimization of attention variants is the natural solution to these problems but the PyTorch compiler stack currently lacks key optimizations such as reduction fusion and complex operator fusion across memory boundaries, limiting the performance of the generated code. 

{\em In this paper, we propose \flashlight{}, a compiler-native mechanism within the PyTorch ecosystem that automatically generates FlashAttention-style tiled and fused kernels from standard PyTorch code without relying on static templates or specialized kernels.}

\flashlight{} turns kernel optimization for attention variants from a manual engineering effort into a \emph{compiler optimization problem}, enabling AI scientists to develop new variants of attention without having to sacrifice performance or scalability. Rather than requiring programmers to re-express their models in a specialized API, \flashlight{} directly analyzes and transforms standard PyTorch attention code, automatically discovering and fusing the constituent operations into a single, tiled Triton kernel. 
It achieves FlashAttention-level performance while preserving the full flexibility of PyTorch.
Users enable \flashlight{} by compiling their PyTorch code using \texttt{torch.compile}. 
\flashlight{} integrates seamlessly with PyTorch’s compiler infrastructure, extending the TorchInductor IR by (1) adopting a unified reduction IR, which enables transforming matrix multiplications; 
(2) captures the algebraic semantics of reductions to enable fusing complex reductions like \textit{softmax}; and 
(3) introduces logical grid dimensions, which enables fusing tiled dimensions. 
\flashlight{} introduces a set of global graph rewrites: 
(1) structural fusion with dimension demotion, 
(2) semantic fusion with algebraic transformation, 
and (3) structural fusion with tiling-aware dimension elimination.
These rewrites can be applied in any order and in combination with existing TorchInductor passes.
Thus, \flashlight{} maintains compatibility as new attention mechanisms and backend optimizations evolve. 

We evaluate several attention variants, including those that are not supported by FlexAttention's template, on H100 and A100 GPUs. 
For variants that FlexAttention supports,
\flashlight{}-generated code is competitive with or faster than that of FlexAttention. 
For all variants, \flashlight{}-generated code is significantly faster than that of torch.compile.
For AlphaFold~\cite{alpafold}, \flashlight{} improves the execution time of row/column-wise gated self-attention by more than $5\times$ and improves the inference latency by $6\%$ to $9\%$.

In summary, \flashlight{} makes the following contributions.
\vspace*{-0.1in}
\begin{closeitemize}
    \item \flashlight{} implements kernel-level fusion that fuses compatible compute blocks (e.g., matmul + softmax, matmul + matmul) when parallelism permits, automatically exploiting opportunities for performance gains.
    \item \flashlight{} achieves generality by supporting both standard and emerging attention mechanisms through a unified, compiler-driven approach rather than relying on static templates or custom kernels.
    \item \flashlight{} produces kernels with competitive or superior performance to FlexAttention, while offering the flexibility of native PyTorch code, enabling developers to rapidly explore new attention models without sacrificing performance.
\end{closeitemize}

\section{Background}\label{sec:background}

This section explains the math behind softmax and its variants, and describes the PyTorch 2.0 compiler stack.


\subsection{Softmax, Safe Softmax, Online Softmax}\label{sec:softmax}

\begin{algorithm}[t]
\caption{Stable softmax.}
\label{alg:stable_softmax}
\begin{algorithmic}[1]
\STATE $m_0\gets -\infty$
\FOR{$k= 1$ to $\vVecLen$}
\STATE $m_k\gets \mathrm{maximum}(m_{k-1},{x_k})$
\ENDFOR
\STATE $d_0\gets 0$
\FOR{$j= 1$ to $\vVecLen$}
\STATE $d_j\gets d_{j-1}+e^{x_j-m_N}$ \label{line:sum_step}
\ENDFOR
\STATE \textbf{Assert: }{$m_\vVecLen = \max{\vVec}$ and $d_N = \sum_{j=1}^\vVecLen e^{x_j - \max{\vVec}}$}
\end{algorithmic}
\end{algorithm}

\begin{algorithm}[t]
\caption{Online softmax}
\label{alg:online_softmax}  
\begin{algorithmic}[1]
\STATE $m_0\gets -\infty$
\STATE $d_0\gets 0$
\FOR{$j\gets 1$ to $\vVecLen$}
\STATE $m_j\gets \mathrm{maximum}\left(m_{j-1},x_j\right)$
\STATE $d_j\gets d_{j-1}\times e^{m_{j-1}-m_j} +e^{x_j-m_j}$\label{line:softmax_restore}
\ENDFOR
\STATE
\STATE \textbf{Assert: }{\(m_\vVecLen = \max{\vVec}\) and \(d_N = \sum_{j=1}^\vVecLen e^{x_j - \max{\vVec}}\)}
\end{algorithmic}
\end{algorithm}

The \emph{softmax} function takes as input a vector 
\(
\vVec = (x_1, x_2, \ldots, x_\vVecLen) \in \mathbb{R}^\vVecLen
\)
and returns a normalized vector \(\sigma(\vVec) \in \mathbb{R}^\vVecLen\) over the \(\vVecLen\) components.
Formally, the \(i\)-th component of the softmax function 
\(\sigma(\vVec)_i\) is defined as:
\begin{equation}
\label{eq:naive_softmax}
\sigma(\vVec)_i 
= \frac{e^{x_i}}{\sum_{j=1}^\vVecLen e^{x_j}} 
\quad
\text{for } i = 1, \ldots, \vVecLen.
\end{equation}
Here, \(e^{x_i}\) denotes exponentiation. 
Eq.~\ref{eq:naive_softmax} ensures that each output 
\(\sigma(\vVec)_i > 0\) and that \(\sum_{i=1}^\vVecLen \sigma(\vVec)_i=1\).
Since exponentiation of large numbers causes numerical instability, a stable implementation uses the maximum of all input elements, $\max \vVec$, as:
\begin{equation}
\label{eq:stable_softmax}
\vSoftmax(\vVec)_i 
= \frac{e^{x_i - \max{\vVec}}}{\sum_{j=1}^\vVecLen e^{x_j - \max{\vVec}}} 
\quad
\text{for } i = 1, \ldots, \vVecLen.
\end{equation}

When computing the denominator of Eq.~\ref{eq:stable_softmax}, a naive implementation uses two serial loops to compute the \texttt{max} and \texttt{sum}.
This implementation is inefficient for large \(\vVecLen\) because it prevents efficient tiling of the computation.


\cite{milakov2018onlinesoftmax} proposed the \textit{online softmax} algorithm, shown in Alg.~\ref{alg:online_softmax}, which fuses the two reduction loops into one by computing the maximum and the normalization denominator simultaneously. This improves performances by reducing the number of memory accesses as the online softmax denominator computation accesses element of the vector $\vVec$ only once. The fused loop also enables tiling and kernel fusion optimizations, which are key to FlashAttention~\cite{dao2023flashattention2fasterattentionbetter}.


\subsection{Attention, FlashAttention, FlexAttention}\label{sec:attention}

Given a sequence of length $n$ represented by \textit{query}, \textit{key}, and \textit{value} matrices $Q, K, V \in \mathbb{R}^{n \times d_k}$, the scaled dot-product attention \cite{vaswani2017attention} is computed as follows:

\vspace*{-0.1in} 
\begin{equation}
\label{eq:attention}
\mathrm{Attention}(Q, K, V) = \mathrm{softmax}\left(\frac{QK^\top}{\sqrt{d_k}}\right) V
\end{equation}
\vspace*{-0.1in}

Softmax is applied row-wise to the attention matrix $\frac{QK^\top}{\sqrt{d_k}}$, producing normalized attention weights that sum to $1$ over the key dimension for each query. 
Listing~\ref{lst:attn} shows the PyTorch implementation of scaled dot-product attention (with \texttt{attn\_mask=None}).
We refer to this as \emph{Vanilla Attention}. 
Computing the attention matrix or \texttt{attn\_scores} requires $\mathcal{O}(n^2)$ time and memory, which becomes a bottleneck for long sequences.

FlashAttention~\cite{dong2024flexattentionprogrammingmodel}, tiles the computation and applies the \textit{online softmax} within each tile.
This design keeps intermediate results in the fast on-chip memory, thus reducing the required memory bandwidth and lowering the memory complexity to $\mathcal{O}(n)$, while maintaining the exact attention computation (in real numbers).
FlashAttention is a hand-optimized kernel for vanilla attention 
and does not support other attention variants.


\lstset{
  language=Python,
  basicstyle=\ttfamily\scriptsize,
  keywordstyle=\color{blue},
  commentstyle=\color{green!50!black},
  stringstyle=\color{red},
  numbers=left,
  numberstyle=\tiny\color{gray}\withcolon,
  frame=single,
  framerule=0.2pt,
  framesep=2pt,
  xleftmargin=20pt,
  xrightmargin=1pt,
  aboveskip=4pt,
  belowskip=4pt,
  breaklines=true,
  breakatwhitespace=true,
  postbreak=\mbox{\textcolor{red}{$\hookrightarrow$}\space},
  columns=flexible,
  keepspaces=true,
  linewidth=\columnwidth
}

\begin{figure}[t]
\begin{lstlisting}[caption={Attention in PyTorch.}, captionpos=b, label={lst:attn}]
def attention(q, k, v, attn_mask=None):
    # Compute attention scores
    attn_scores = torch.matmul(q, k.transpose(-2, -1))
    attn_scores *= 1 / math.sqrt(q.size(-1))
    
    if attn_mask is not None: 
        # Set -INF when mask is True
        attn_scores = attn_scores.masked_fill(
                        attn_mask, -INF)
    
    # Apply softmax and compute output
    attn_weights = torch.softmax(attn_scores, dim=-1)
    output = torch.matmul(attn_weights, v)
    return output
\end{lstlisting}
\end{figure}



FlexAttention~\cite{dong2024flexattentionprogrammingmodel} provides a template to write attention variants that modify the attention score:
\vspace*{-0.05in}
\begin{equation}
\begin{split} 
\mathrm{FlexAttention}&(Q, K, V, score\_mod) = \\
&\mathrm{softmax}\left(\mathrm{score\_mod}\left(\frac{QK^\top}{\sqrt{d_k}}\right)\right) V
\end{split}
\label{eq:flex-attn}
\end{equation}


\lstset{
  language=Python,
  basicstyle=\ttfamily\scriptsize,
  keywordstyle=\color{blue},
  commentstyle=\color{green!50!black},
  stringstyle=\color{red},
  numbers=left,
  numberstyle=\tiny\color{gray}\withcolon,
  frame=single,
  framerule=0.2pt,
  framesep=2pt,
  xleftmargin=20pt,
  xrightmargin=2pt,
  aboveskip=4pt,
  belowskip=4pt,
  breaklines=true,
  breakatwhitespace=true,
  postbreak=\mbox{\textcolor{red}{$\hookrightarrow$}\space},
  columns=flexible,
  keepspaces=true,
  linewidth=\columnwidth,
  escapechar=|
}

\begin{figure}[t]
\begin{lstlisting}[caption={Sliding Window Attention using FlexAttention.}, captionpos=b, label={lst:flex-sliding-window}]
from torch.nn.attention.flex_attention import (
        flex_attention, create_block_mask)
from functools import lru_cache

@lru_cache |\label{line:flex-attn:sliding-window:cache}|
def create_block_mask_cached(mask, B,H,M,N, device):
    return create_block_mask(mask, B,H,M,N, device)

def generate_sliding_window(window_size):
    def sliding_window_mask(b, h, q, kv): |\label{line:flex-attn:sliding-window:mask}|
        return (q >= kv) & ((q - kv) <= window_size)
    # Set -INF when mask is False
    return sliding_window_mask

# Create block mask for sliding window
block_mask = create_block_mask_cached(|\label{line:flex-attn:sliding-window:block-mask}|
    generate_sliding_window(window_size=256),
    B=1, H=1, # ignore actual B and H sizes
    M=q.size(2), N=k.size(2), device="cuda")

# Compile for performance
flex_attn = torch.compile(flex_attention, 
                        dynamic=False)

# Apply attention with sliding window's block mask
output = flex_attn(q, k, v, block_mask=block_mask)
\end{lstlisting}
\end{figure}

FlexAttention only supports attention variants that fit this pattern. 
Users specify the \texttt{score\_mod} as an element-wise operation on the attention matrix that accepts an old score and returns a new score. 
For performance reasons, they introduce a special case for \texttt{score\_mod} called \texttt{mask\_mod}. 
Users specify whether a given index of the attention matrix should be masked (score set to infinity) or not (use old score). 
As this function only depends on the shape of $Q$ and $K$, they further specialize for this case by providing a PyTorch function 
that inspects the \texttt{mask\_mod} at runtime to create a custom \texttt{block\_mask} representation that stores (in device memory) sparse matrices for empty, full, or partial blocks.
FlexAttention uses the PyTorch compiler stack to generate a fused and optimized kernel from a specialized template that executes attention with these inspected sparse matrices.
For example in Listing~\ref{lst:flex-sliding-window}, the \emph{sliding window attention}~\cite{beltagy2020longformerlongdocumenttransformer} applies a sliding window mask (Line~\ref{line:flex-attn:sliding-window:mask}) to only consider the attention scores within a window.
For performance, users are expected to both create the block mask (Line~\ref{line:flex-attn:sliding-window:block-mask}) to represent this window and cache this mask (Line~\ref{line:flex-attn:sliding-window:cache}) to avoid its recomputation on later calls.



\begin{figure*}[h!]
    \centering
    \includegraphics[width=\textwidth]{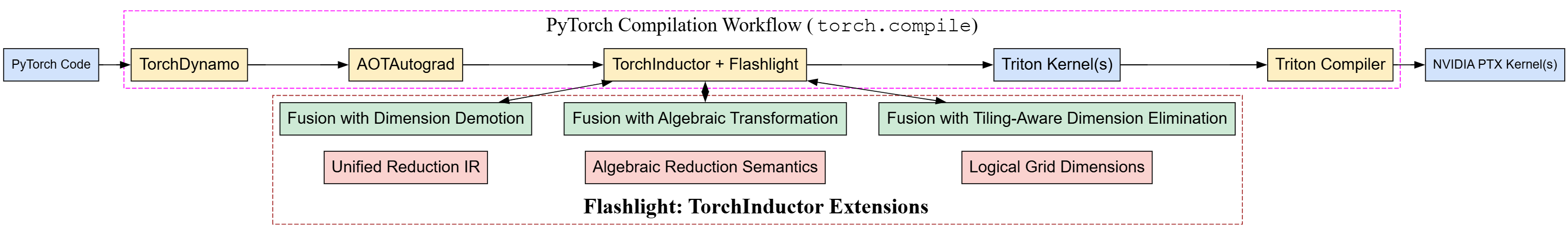}
    \caption{\flashlight{} extends TorchInductor within the \texttt{torch.compile} stack, adding structural and semantic fusion passes with dimension demotion, algebraic transformation, and tiling-aware dimension elimination to generate optimized Triton kernels.}
    \label{fig:flashlight_overview}
\end{figure*}

\subsection{The PyTorch 2.0 Compiler Stack}\label{sec:torchcompiler}

\newcommand{\backgroundtrim}[1]{\revised{#1}}

\backgroundtrim{PyTorch is widely adopted for machine learning due to its eager execution model, which treats model definitions as imperative Python code. This design makes PyTorch flexible and intuitive but complicates compiler optimizations that rely on static, graph-based representations of computation. 
Unlike frameworks like TensorFlow~\cite{Abadi2016tensorflow} or Theano~\cite{Bergstra2010theano},}
PyTorch does not natively expose a full program graph suitable for whole-graph analysis and transformation. The \emph{PyTorch 2.0 compiler stack}~\cite{Ansel2024pytorch2}, used via the \verb|torch.compile()| API, addresses this limitation 
by using two key components: \emph{TorchDynamo}, a Python-level graph extractor, and \emph{TorchInductor}, a backend compiler targeting both CPUs and GPUs.

{\bf TorchDynamo} extracts the program graph by detecting  
PyTorch operations, which are transformed into an FX graph~\cite{Reed2022torchfx}, an intermediate representation designed for further optimization and lowering. 
TorchDynamo integrates with AOTAutograd, which enables training by recording both forward and backward graphs. 

{\bf TorchInductor} functions as a general-purpose compiler backend designed to translate FX graphs into optimized, high-performance code. 
\backgroundtrim{It supports multiple backends: a built-in OpenAI Triton~\cite{Tillet2019Triton} for GPU, C++ with OpenMP for CPU execution, and custom backends defined by the user.} 
TorchInductor introduces a Python-based loop-level intermediate representation (IR) using a define-by-run model: tensor computations are expressed as Python functions over symbolic indices.
This IR enables (i) \textit{operator decomposition} into a minimal core set of pointwise, reduction, and other primitives operations, (ii) \textit{fusion and scheduling} informed by symbolic analysis of memory access patterns and aliasing, and (iii) \textit{efficient code generation} with optional use of CUDA Graphs to minimize kernel launch overhead.
TorchInductor supports vectorization, auto-tuning, and ahead-of-time kernel compilation. 

{\bf Dynamic Shapes} are supported in the PyTorch~2.0+ compiler stack. Using meta-functions that track and propagate shape information symbolically, TorchDynamo and TorchInductor can reuse compiled code for inputs of different sizes. The compiler also uses guards and simple symbolic checks to decide when it needs to recompile. For models with fixed input sizes, users set \verb|dynamic=False| in \verb|torch.compile()| to turn off dynamic shape tracing and generate shape-efficient kernels for faster performance.

\backgroundtrim{
{\bf Performance} PyTorch 2.0 achieves strong performance on real-world workloads. Across 180+ models from TorchBench, HuggingFace, and TIMM, TorchDynamo and TorchInductor deliver a geometric mean speedup of $2.27\times$ on inference and $1.41\times$ on training with float32 on NVIDIA A100 GPUs~\cite{Ansel2024pytorch2}. TorchDynamo’s graph capture overhead is under 5\%, significantly outperforming Lazy Tensors and previous JIT mechanisms.}

\backgroundtrim{
The PyTorch 2.0 compiler stack introduces a hybrid approach to dynamic language compilation, combining bytecode-level tracing with backend-level optimization and code generation. It makes aggressive compiler optimization accessible to users without compromising Python’s dynamism or usability.}

\section{Flashlight}
\label{sec:flashlight}


\lstset{
  language=Python,
  basicstyle=\ttfamily\scriptsize,
  keywordstyle=\color{blue},
  commentstyle=\color{green!50!black},
  stringstyle=\color{red},
  numbers=left,
  numberstyle=\tiny\color{gray}\withcolon,
  frame=single,
  framerule=0.2pt,
  framesep=2pt,
  xleftmargin=20pt,
  xrightmargin=2pt,
  aboveskip=4pt,
  belowskip=4pt,
  breaklines=true,
  breakatwhitespace=true,
  postbreak=\mbox{\textcolor{red}{$\hookrightarrow$}\space},
  columns=flexible,
  keepspaces=true,
  linewidth=\columnwidth
}

\begin{figure}[t]
\begin{lstlisting}[caption={Sliding Window Attention using Flashlight.}, captionpos=b, label={lst:flashlight-sliding-window}]
def get_sliding_mask(query, window):
    L = query.size(-2)
    idx = torch.arange(L, device=query.device)
    q = idx.view(L, 1)
    k = idx.view(1, L)
    mask = (q < kv) or ((q - kv) > window) #(L, L)
    return mask.unsqueeze(0).unsqueeze(0)

def sliding_window_attn(q, k, v, window):
    attn_mask = get_sliding_mask(q, window)
    return attention(q, k, v, attn_mask=attn_mask)

# Compile with Flashlight enabled
sliding_window_attn = torch.compile(
                sliding_window_attn, dynamic=False, 
                enable_flashlight=True)

output = sliding_window_attn(q, k, v, window=256)
\end{lstlisting}
\end{figure}

\begin{figure}[t]
\begin{lstlisting}[caption={Differential Attention using Flashlight.}, captionpos=b, label={lst:diffattn}]
def diff_attn(q, k, v, lambda_full):
    # Split Q and K into two heads
    q0, q1 = q.chunk(2, dim=1)
    k0, k1 = k.chunk(2, dim=1)
    
    # Compute attention for both heads
    attn0 = attention(q0, k0, v)
    attn1 = attention(q1, k1, v)
    
    # Differential: subtract weighted second head
    output = attn0 - lambda_full * attn1
    return output

# Compile with Flashlight enabled
diff_attn = torch.compile(
                diff_attn, dynamic=False,
                enable_flashlight=True)

output = diff_attn(q, k, v, lambda_full=0.2)
\end{lstlisting}
\end{figure}

Figure~\ref{fig:flashlight_overview} illustrates the \texttt{torch.compile} workflow and how \flashlight{} plugs into it.
\flashlight{} extends TorchInductor with a principled approach to operator fusion and scheduling, based on a set of composable compiler transformations.
This allow it to dynamically fuse complex subgraphs, such as the various forms of attention, into monolithic, high-performance kernels without requiring explicit user annotations or modifications to idiomatic PyTorch code. 
Listing~\ref{lst:flashlight-sliding-window} shows \emph{sliding window attention}~\cite{beltagy2020longformerlongdocumenttransformer} using \flashlight{}. 
It uses idiomatic PyTorch code 
without requiring the user to build a \texttt{block\_mask} or use a cache like in Listing~\ref{lst:flex-sliding-window} for FlexAttention.
A user just needs to pass a flag to \texttt{torch.compile} to enable \flashlight{}.
This approach stands in contrast to existing systems that often depend on predefined static code templates (e.g., TorchInductor replaces vanilla attention with a hand-optimized kernel by searching for patterns in its IR) or explicit user annotations (e.g., FlexAttention), limiting their applicability beyond common patterns like Equation~\ref{eq:flex-attn}.
As a consequence, \flashlight{} enables compiling complex attention patterns that are not expressible in FlexAttention, such as differential attention~\cite{ye2024differentialtransformer} (Listing~\ref{lst:diffattn}), row/column-wise gated self-attention in Evoformer~\cite{alpafold}, IPA~\cite{alpafold}, and RSA~\cite{rectifiedsparseattention}.

We demonstrate Flashlight's approach using the vanilla attention in Eq~\ref{eq:attention}.
\flashlight{} automatically fuses the entire sequence of vanilla attention in Listing~\ref{lst:attn} into a single monolithic Triton kernel.
This kernel executes the computation in a single pass over the input data, where each thread block computes tiles of the dot-product $S = QK^\top/\sqrt{d}$, applies the "online" softmax to each tile via a fused max-reduction and rescaled accumulation, and multiplies the resulting softmax output with the corresponding tiles of $V$.

Section~\ref{subsec:unified-ir} describes the unified reduction IR in TorchInductor that enables expressing the loops of matrix multiplications like $QK^\top$. 
Section~\ref{subsec:fusion-dim-demotion} describes structural fusion of matrix multiplications with simple reductions by demoting dimensions to fuse the $max()$ operation in $softmax()$ with the preceding $QK^\top$. 
Section~\ref{subsec:alg-reduction} describes the semantics of algebraic transformation of reductions that generalizes the transformation of the stable softmax algorithm to the online softmax algorithm. 
We use this in TorchInductor to semantically fuse operations like $\mathrm{softmax}\bigl(QK^\top/\sqrt{d}\bigr)$ in Section~\ref{subsec:fusion-alg-reduction}.
Section~\ref{subsec:fusion-tiling} describes structural fusion of tiled loops while eliminating small dimensions and 
Section~\ref{subsec:logical-grid} describes ways to accomplish this in TorchInductor using logical grid dimensions. 
This would fuse consecutive matrix multiplication operations like $\mathrm{softmax}\bigl(QK^\top/\sqrt{d}\bigr)V$.
\revised{We discuss some implementation details specific to \texttt{torch.compile} in Section~\ref{subsec:pytorch-impl-detail}.}
Finally, we discuss the generality, flexibility, and trade-offs of Flashlight and compare it with FlexAttention 
(Section~\ref{subsec:trade-offs}).

\subsection{Unified Reduction IR}
\label{subsec:unified-ir}

To provide a unified intermediate representation (IR) for tensor computations, tensor dimensions are classified into two categories: \textit{p}-dimensions (\textbf{p}arallel/\textbf{p}ointwise) and \textit{r}-dimensions (\textbf{r}eduction).
Computations over p-dimensions are data-independent, representing embarrassingly parallel workloads that can be mapped directly to parallel execution units, such as programs in the Triton language or CUDA thread blocks.
For instance, in an element-wise operation like \texttt{torch.add(A, B)}, all dimensions of tensors A and B are p-dimensions.
Conversely, r-dimensions introduce data dependencies that require sequential accumulation or specialized parallel primitives like reduction trees.
A canonical example is \texttt{torch.sum(T, dim=n)}, where dimension $n$ is an r-dimension, as its elements are aggregated, while all other dimensions remain p-dimensions.

This conceptual framework is operationalized by TorchInductor, where it lowers operations into a loop-level IR that explicitly categorizes them based on their dimensional properties.
However, TorchInductor has a special path for performance-sensitive tensor contractions like General Matrix Multiplication (GEMM), where it bypasses the IR-to-kernel generation path to either instantiate a pre-written, highly-optimized kernel template of the Triton language or generate calls to vendor-tuned libraries like ATen or cuBLAS.
While this workflow ensures high performance for the standalone GEMM operation, this bifurcation creates a \textit{fusion boundary} that isolates the GEMM from surrounding computations.
This prevents fusion of GEMM with preceding or subsequent operations beyond the support for simple element-wise operations,
reintroducing  memory bandwidth bottlenecks and kernel launch overheads.

\flashlight{} dismantles this artificial fusion boundary by modeling tensor contractions as a generalized reduction within our unified IR, which is compatible to the PyTorch built-in reduction IR. We observe that a GEMM operation naturally conforms to the p- and r-dimension abstraction.
Consider the canonical 2D matrix multiplication, $C = A \cdot B$, defined as $C_{mn} = \sum_k A_{mk} B_{kn}$. We model this as follows:
\vspace*{-0.1in}
\begin{closeitemize}
    \item The output dimensions, indexed by $m$ and $n$, are \textbf{p-dimensions}. The computation for each output element $C_{mn}$ is independent of all others, and these dimensions are preserved in the output tensor's shape.
    \item The inner contracted dimension, indexed by $k$, is an \textbf{r-dimension}. It is iterated over, and the products are accumulated via a \textit{sum} reduction. This dimension is shared by both inputs and consumed by the operation and does not appear in the output.
\end{closeitemize}
By representing GEMM within the same semantic framework as other tensor operations, \flashlight{} enables it to participate fully in our end-to-end fusion engine. This unified approach unlocks advanced optimizations, such as fusing chains of matrix multiplications or complex element-wise prologues directly into a single, efficient kernel.

\subsection{Structural Fusion with Dimension Demotion}
\label{subsec:fusion-dim-demotion}

The p- and r-dimension abstraction provides a formal basis for a canonical loop structure.
For a given tensor operation, the data-independent p-dimensions form the outer parallel loops, while the data-dependent r-dimensions form the inner iterative loops.
This structure naturally maps to the hierarchical execution model of modern GPUs, where outer loops can be parallelized across thread blocks, and inner loops are executed sequentially across warps.

We can therefore define a \textbf{computation sketch} for a kernel, denoted as $[(P_0, P_1, \dots), (R_0, R_1, \dots)]$, which captures its loop hierarchy.
{\scriptsize
\begin{verbatim}
for p0 in P0
  for p1 in P1
    ...
    for r0 in R0
      for r1 in R1
      ...
\end{verbatim}
}
where the capital letters are the iteration ranges of dimensions 
and the lowercase letters are used for indexing.For example:

\begin{closeitemize}
    \item \textbf{Element-wise Addition}: $C(P_0, P_1) = A(P_0, P_1) + B(P_0, P_1)$ involves no reduction and has the sketch $[(P_0, P_1), ()]$. Since the p-dimensions are data-independent, the computation can be perfectly flattened and parallelized across all elements.

    \item \textbf{GEMM}: $C(P_0, P_1) = A(P_0, R_0) @ B(R_0, P_1)$ involves a sum-reduction over the inner dimension and has the sketch $[(P_0, P_1), (R_0)]$. The computation for each output element $(p_0, p_1)$ requires an inner loop over $r_0$.
\end{closeitemize}

Under this model, existing compilers like TorchInductor primarily fuse operations with identical sketches. For instance, two kernels with the sketch $[(P_0, P_1), ()]$ can be fused vertically.
Fusion is also permitted between a pointwise operation and a reduction if their p-dimensions align (e.g., fusing $[(P_0, P_1), ()]$ into $[(P_0, P_1), (R_0)]$).
This model, however, is not sufficient for fusing operations where the dimensions fundamentally realign, such as in producer-consumer patterns where the producer's output dimension becomes a reduction dimension for the consumer.

Flashlight introduces a more powerful fusion rule by leveraging a key insight that \textit{a parallel loop can be executed sequentially}.
This allows us to "demote" a p-dimension from a producer kernel into an r-dimension within the fused kernel.
Formally, a producer kernel $K_0$ with sketch $[(P_{common}, P_{producer}), (\dots)]$ can be fused with a consumer kernel $K_1$ with sketch $[(P_{common}), (P_{producer}, \dots)]$. The resulting fused kernel will have the sketch:
$ [(P_{common}), (P_{producer}, \dots)] $.
Here, the dimension $P_{producer}$, which was a p-dimension in the standalone producer kernel $K_0$, becomes an additional inner reduction loop in the fused kernel.

The rationale for this transformation is the fundamental trade-off between parallelism and memory latency. By demoting a parallel dimension, we tradeoff some of the producer's potential parallelism for complete elimination of the high-latency materialization of the intermediate tensor to global memory.
As a consequence, the producer's results are generated and consumed immediately within the registers or local memory of a single, unified kernel.
On modern accelerators where memory bandwidth is often a more critical bottleneck than raw compute parallelism, this trade-off is overwhelmingly favorable, leading to significant performance improvements.
An example for this scenario is fusing only the $max()$ operation inside $softmax()$ with the preceding $QK^\top$ in attention.

\subsection{Algebraic Transformation of Reductions}
\label{subsec:alg-reduction}

The Online softmax algorithm (Alg.~\ref{alg:online_softmax}) is key to implementing FlashAttention-like fused kernel.
However, current compilers cannot generate the online implementation automatically. In Appendix~\ref{sec:alg-reduction-appendix}, we show that the conversion of the stable softmax algorithm to the online softmax algorithm can be generalized using the standard algebraic notion of a {\em homomorphism}. We describe the high level idea here. 

\vspace*{-0.1in}
\begin{definition}
\label{def:homomorphism}
Let $A$ be a set with a binary operation $\oplus$, and let $B$ be a set with a binary operation $\otimes$. A function $f:A{\rightarrow}B$ 
is said to be a homomorphism if for all $a_1,a_2 \in A$, $f(a_1\oplus a_2) = f(a_1)\otimes f(a_2)$. 
\end{definition}
In the context of softmax, $A = B = \mathbb{R}$ (the set of real numbers), and $\oplus$ and  $\otimes$ are addition ($+$)  and multiplication ($\times$) of real numbers. The function $f(x) = e^x$ is a homomorphism because $f^{a+b} = f^a\times f^b$. To generalize the online softmax construction, we need the set $A$ with operations $\oplus$ and $\otimes$ to satisfy the axioms of a {\em ring}.
\vspace*{-0.1in}

In the stable softmax Algorithm~\ref{alg:stable_softmax}, let us denote the sequence of $m$ values produced by the first loop by $m[1..N]$ and let $m[0]=0$ by definition. The sequence of $d$ values produced by the second loop, which we denote by $ds$, is expressed abstractly by the following recurrence in which the elements of $ds$ are members of a ring $A$ and $E{:}A{\rightarrow}A$ is a homomorphism.  

{\scriptsize
\begin{mdframed}
\begin{flalign*}
ds[0] &= 0 &\\
ds[j] 
&= ds[j{-}1] \oplus (E(x[j])\otimes E(\ominus m[N])) \ ~|N\geq j \geq 1 &
\end{flalign*}
\end{mdframed}
}

The online softmax Algorithm~\ref{alg:online_softmax} computes a different sequence, denoted by $do$, that can be expressed abstractly as shown below. 

{\scriptsize
\begin{mdframed}
\vspace*{-0.1in}
\begin{flalign}
do[0] &= 0  \\
\begin{split}
do[j]  &= 
(E(x[j])\otimes E(\ominus m[j])) ~|N\geq j \geq 1
\label{doDef}
\end{split}&
\end{flalign}
\end{mdframed}
}

The $ds$ and $do$ sequences will be different in general, but we show that $do$ can be expressed in closed-form by the following expression.

{\scriptsize
\begin{mdframed}
\vspace*{-0.1in}
\begin{flalign}
do[j]  & = \bigg(\bigoplus_{i{=}1}^j E(x[i])\bigg) \otimes E(\ominus m[j])  \ ~|N \geq j \geq 1 & \label{doClosed}
\end{flalign}
\end{mdframed}
}

from which it follows that $ds[N] = do[N]$. The proof is in the Appendix~\ref{sec:alg-reduction-appendix}.

\subsection{Semantic Fusion with Algebraic Transformation}
\label{subsec:fusion-alg-reduction}

Beyond the structural compatibility addressed by computation sketches, valid kernel fusion must also respect data dependencies.
A challenge for fusion arises from \textit{cross-kernel data dependencies}.
For instance, one reduction kernel depends on the fully aggregated result of another.
This dependency creates a synchronization barrier that prevents naive loop fusion, even if the kernels share an identical sketch. A canonical example of this challenge is the numerically stable softmax operation, which involves two sequential passes for numerical stability: first finding the maximum value of the input tensor, and second, computing the sum of exponentials shifted by that maximum.
\vspace*{-0.1in}
\begin{closeenumerate}
    \item \textbf{Pass 1 (max):} $m_i = \max(m_{i-1}, A_i)$.
    \item \textbf{Pass 2 (sub-exp-sum):} $S = \sum_i \exp(A_i - m_{final})$.
\end{closeenumerate}
Although both operations have a compatible reduction sketch, e.g., $[(), (R)]$, they cannot be trivially fused because Pass 2 has a strict dependency on $m_{final}$ — the final scalar result of the first kernel.
A direct merge by stacking the loop bodies would be incorrect, as the computation of each term $\exp(A_i - m_{final})$ requires a value that is only known after the first loop has fully completed.


\flashlight{} overcomes this barrier by identifying and transforming these dependent computation into a single-pass, online algorithm when an underlying algebraic structure permits it.
The key is to transform the dependency on the \textit{final} result into an incremental update based on the \textit{running} result.
This is possible here due to the \textit{homomorphic} property of the exponential function (Section~\ref{sec:alg-reduction-appendix}), which maps addition/subtraction to multiplication/division: $\exp(x - y) = \exp(x) / \exp(y)$.
This property allows us to dynamically "rescale" the running sum whenever the running maximum changes.

Within a single loop, two running accumulators are maintained: the running max ($m_r$) and a running sum ($s_r$).
The challenge is that whenever the running max is updated, the accumulated sum becomes invalid as it was normalized by a now-outdated maximum.
If the sum was calculated using an old max, $m_{old}$, and the max is updated to $m_{new}$, the corrected sum can be found by multiplying by a correction factor:

{\scriptsize
$$ S_{old} = \sum_i \exp(A_i - m_{old}) = \sum_i \frac{\exp(A_i)}{\exp(m_{old})}$$
}
{\scriptsize
$$ S_{new} = \sum_i \exp(A_i - m_{new}) = \sum_i \frac{\exp(A_i)}{\exp(m_{old})} \frac{\exp(m_{old})}{\exp(m_{new})} = $$
$$S_{old} \times \exp(m_{old} - m_{new}) $$
}

This allows us to fuse the two passes into a single, efficient kernel. In each step of the loop, we update both the running max and the running sum simultaneously. If an element $A_i$ causes the running max to change, we apply the correction factor to the sum accumulated so far before adding the new term.
To accomplish this in TorchInductor, we decompose a reduction to introduce a loop-local variable that copies the partially-aggregated, loop-carried value before applying the aggregation in this step. 

By embedding such algebraic reasoning into TorchInductor, Flashlight can semantically restructure the algorithm to fuse complex, state-dependent operations that are beyond the scope of fusion based on structural compatibility alone.
This unlocks end-to-end kernel fusion for a broader class of complex, multi-stage reductions like $\mathrm{softmax}\bigl(QK^\top/\sqrt{d}\bigr)$.

\subsection{Structural Fusion with Tiling-Aware Dimension Elimination}
\label{subsec:fusion-tiling}

The loop-based execution model can be further refined by iterating over contiguous \textit{tiles} (or \textit{blocks}) of data rather than individual elements.
Practical GPU kernels adopt tiling as a critical optimization that improves data locality by staging data in fast on-chip memory like shared memory or registers.
The tiled execution also structures the computation to better exploit the GPU's SIMT architecture with fine-grained parallelism within each tile.
For instance, an associative reduction over a tile can be implemented efficiently using parallel primitives like warp-level tree reductions or by vectorizing the accumulation across multiple SIMT lanes.

Crucially, tiling transforms the loop structure of a kernel. A loop over a dimension of size $D$ with a tile size of $B_D$ will execute $\lceil D / B_D \rceil$ times at the tile level. This allows us to define a \textit{tiled sketch} for a kernel:
{\scriptsize
$ [(\lceil\frac{P_0}{B_{P0}}\rceil, \dots), (\lceil\frac{R_0}{B_{R0}}\rceil, \dots)] $}

This transformation from an element-space sketch to a tile-space sketch creates more opportunities for fusion.
The key insight is that, if a dimension $P_i$ is small enough to be processed entirely within a single tile (i.e., $B_{Pi} \geq |P_i|$), the corresponding tile-level loop bound becomes $\lceil |P_i| / B_{Pi} \rceil = 1$.
A loop with a single iteration can be conceptually elided from the sketch, effectively collapsing the dimension at the tile level.
This tiling-aware dimension elimination unlocks fusion opportunities that are otherwise impossible.

Consider  twin matrix multiplication, $E = (A \cdot B) \cdot D$.
\begin{closeitemize}
    \item \textbf{Kernel 0 (Producer):} {\scriptsize$C[M, N] = A[M, K] @ B[K, N]$}. The element-space sketch is {\scriptsize$[(M, N), (K)]$}.
    \item \textbf{Kernel 1 (Consumer):} {\scriptsize$E[M, P] = C[M, N] @ D[N, P]$}. The element-space sketch is {\scriptsize$[(M, P), (N)]$}.
\end{closeitemize}
These kernels are incompatible under standard fusion rules. Without tiling, these sketches are also incompatible for the producer-consumer fusion described previously; the producer's p-dimension $N$ does not match the consumer's p-dimension $P$. Fusing them would require materializing the entire intermediate tensor $C$ into global memory.

\flashlight{} can fuse these kernels by leveraging tiling.
Let's assume the dimension $P$ is relatively small (e.g., 64, 128).
We can set the tile size for the $P$ dimension in Kernel 1 to be its full size, $B_P = |P|$.
The tiled sketch for Kernel 1 becomes:

{\scriptsize
$$ [(\lceil\frac{M}{B_M}\rceil, \lceil\frac{P}{B_P}\rceil=1), (\lceil\frac{N}{B_N}\rceil)] \implies [(\lceil\frac{M}{B_M}\rceil), (\lceil\frac{N}{B_N}\rceil)] $$
}

The $P$ dimension has been collapsed out of the consumer's tile-level p-dimensions.
Now, the producer sketch
{\scriptsize$[(M_{tile}, N_{tile}), (K_{tile})]$}
and the consumer sketch
{\scriptsize$[(M_{tile}), (N_{tile})] $}
can be combined as
{\scriptsize$[(M_{tile}), (N_{tile}, K_{tile})]$}

using our generalized fusion rule with dimension demotion.
The dimension $N$ from the producer's output is consumed directly as an r-dimension by the consumer on-the-fly within a single fused kernel, completely avoiding the materialization of the intermediate tensor $C$.
This tiling-aware transformation allows \flashlight{} to fuse complex, multi-stage computations that have historically remained out of reach for the PyTorch compiler.

\subsection{Flexible Tiling through Logical Grid Dimensions}
\label{subsec:logical-grid}


Practical implementations of tiling strategies in TorchInductor must address some framework constraints.
The ideal approach allows for flexible, per-dimension tile sizes, enabling a broad autotuning search space and the dimension elimination technique.
However, systems like TorchInductor often create a rigid coupling between the logical tiling dimensions and the physical GPU execution grid.
This coupling arises from the asymmetric hardware limits of the grid; in CUDA, for example, the X dimension can be up to $2^{31}-1$, while the Y and Z dimensions are limited to $65,535$.
To accommodate large tensors, TorchInductor often flattens multiple p-dimensions into the expansive X grid dimension.
This presents a dilemma:
\vspace*{-0.1in}
\begin{closeitemize}
    \item Flattening: For a computation with a sketch $[(P_0, P_1)]$, both dimensions are mapped to the X grid. TorchInductor forces them to share a single tile size, preventing independent tuning of the tile sizes for $P_0$ and $P_1$.
    \item Multi-Grid Mapping: Mapping $P_0$ to the Y grid and $P_1$ to the X grid would allow separate tile sizes but this approach fails if the cardinality of dimension $P_0$ exceeds $65,535$.
\end{closeitemize}
This inflexibility restricts the compiler's ability to find the optimal tiling configuration.

\flashlight{} resolves this dilemma by decoupling the problem's structure from the hardware's using \textit{logical tiling}.
Instead of directly affine mapping tiling dimensions to physical grid dimensions, \flashlight{} defines a logical, multi-dimensional grid of tiles based on the per-dimension tile sizes.
This logical grid is then "unrolled" into a linear sequence and mapped to a single physical grid dimension (e.g., \texttt{tl.program\_id(0)} in Triton).
Inside the kernel, a simple inverse affine map recovers the logical multi-dimensional tile coordinates from the linear block ID.

This approach 
allows tile sizes for each dimension to be set independently by the autotuner or controlled via hints.
This ensures that powerful fusion strategies, like the tiling-aware dimension elimination, can be applied robustly across a far wider range of tensor shapes and sizes.

\subsection{Implementation Details}
\label{subsec:pytorch-impl-detail}

\paragraph{Precision handling.}
\revised{
When GEMM is lowered as a generalized reduction in our unified IR, the computation type is unconditionally promoted to FP32 for FP16/BF16 inputs. This ensures numerical stability during accumulation (matching the behavior of hardware tensor cores) while keeping the input/output in the original lower precision. The BF16 output type is preserved through the decomposition and meta-registration layers, with an ad-hoc type restore in the graph lowering to prevent the compiler from inadvertently widening the output type.
}

\paragraph{L2 cache optimization.}
\revised{
For kernels with two or more tiled parallel dimensions, the block iteration order is swizzled to improve spatial locality. Blocks are grouped into strips of width \texttt{GROUP\_M}, and within each strip, the iteration alternates between dimensions to maximize L2 cache reuse for adjacent tiles. This is analogous to the swizzling technique used in Triton's matmul tutorial~\footnote{https://triton-lang.org/main/getting-started/tutorials/03-matrix-multiplication.html\#l2-cache-optimizations}, generalized to arbitrary multi-dimensional tiling.
}

\paragraph{Indexing order tracking.}
\revised{
When the compiler's symbolic engine (e.g., Sympy) rewrites an index expression like $m \cdot N + n$ into a simplified form, the variable order may be lost.
\flashlight{} records the ordering of index variables as they flow through symbolic simplifications and loop-body substitutions.
The record preserves this mapping so that the codegen phase can emit correctly-shaped N-dimensional tensors, masks, and loads even after aggressive index rewriting.
}

\paragraph{Block-reduction heuristic.}
\revised{
\flashlight{} introduces a new kernel launch heuristic, \texttt{blockreduction}, for fused block-reduction kernels. It uses template-based autotuning over a configuration space of $(\texttt{XBLOCK}, \texttt{RBLOCK}, \texttt{num\_warps}, \texttt{num\_stages})$ tuples.
With aggressive autotuning enabled, the search space expands to more configurations including smaller block sizes for workloads with limited parallelism. When the scheduler provides block-size hints (from the blocking analysis), these override the default search space to focus autotuning on the most promising region.
}

\paragraph{Materialization threshold.}
\revised{
The baseline compiler's defines a materialization threshold, the maximum number of fused operations before an intermediate tensor is forced to be materialized to global memory.
\flashlight{} raises the threshold to allow more complex fused subgraphs (such as \emph{ALiBi}~\cite{press2022trainshorttestlongalibi}) to remain in a single kernel without premature materialization.
}

\subsection{Generality, Flexibility, Tradeoffs}
\label{subsec:trade-offs}

The transformations in \flashlight{} are general and widely applicable, but they involve performance and accuracy trade-offs. 
Structural fusion (Section~\ref{subsec:fusion-dim-demotion}) reduces memory accesses at the cost of reduced parallelism. 
Semantic fusion (Section~\ref{subsec:fusion-alg-reduction}) preserves algebraic semantics in real numbers, but may not be precise in floating point numbers because floating point arithmetic is not associative. 
While these transformations optimize a variety of attention variants without observable loss in model accuracy~\cite{dao2022flashattentionfastmemoryefficientexact}, they may not improve performance or may be inaccurate for some kernels on some hardware. 
Users should enable \flashlight{} for PyTorch code based on this knowledge.

\flashlight{} enables users to explore a wide variety of attention variants and other tensor computations without having to map their code to specific patterns or sacrifice performance or scalability. 
This flexibility also has performance trade-offs. 
Unlike FlexAttention or other attention-specific DSLs,
\flashlight{} does not specialize its transformations for the mask in certain attentions. 
FlexAttention, for example, inspects the mask to build a \texttt{block\_mask} of sparse matrices using which it can skip attention for the tiles or blocks that are fully masked out. 
When the matrices are very sparse, the attention execution can be much faster 
because it reduces redundant computation. However, the inspection code to create \texttt{block\_mask} is expensive and storing the \texttt{block\_mask} consumes resource-constrained GPU memory. 
Users are expected to manage this trade-off by building a cache and managing its size. 
Moreover, even with a cache hit, executing attention with inspected sparse matrices might have runtime overheads. 
Thus, such inspection-execution code is not always beneficial.
We leave incorporation of such techniques in \texttt{torch.compile()} to future work.

\section{Evaluation}
\label{sec:eval}

\begin{figure*}
\begin{minipage}{\textwidth}
  \centering
  \includegraphics[width=0.95\textwidth]{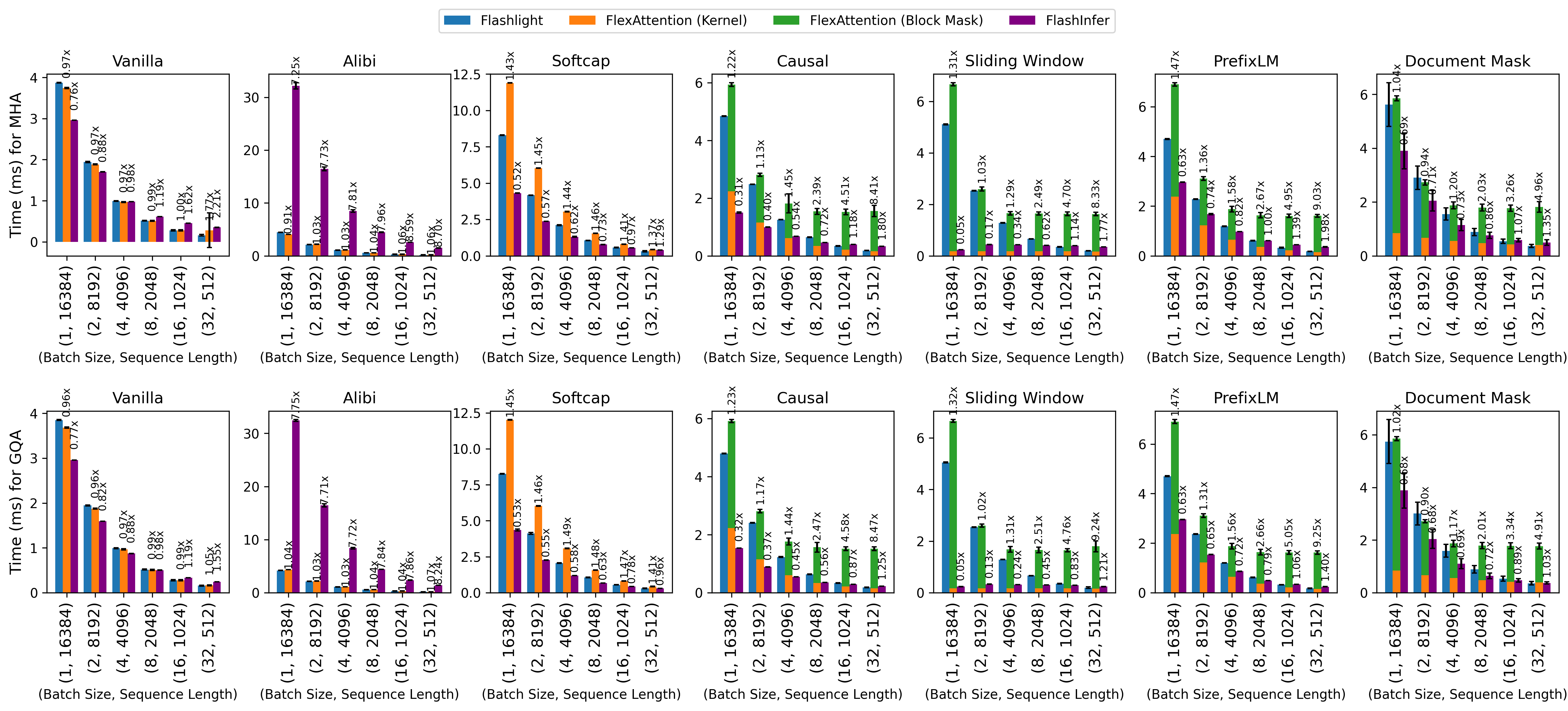}
  \caption{\revised{\flashlight{}, FlexAttention, and FlashInfer} on H100 for attention variants that are supported by FlexAttention template.}
  \label{fig:flex-able-h100}
\end{minipage}\\
\begin{minipage}{\textwidth}
  \centering
  \includegraphics[width=0.95\textwidth]{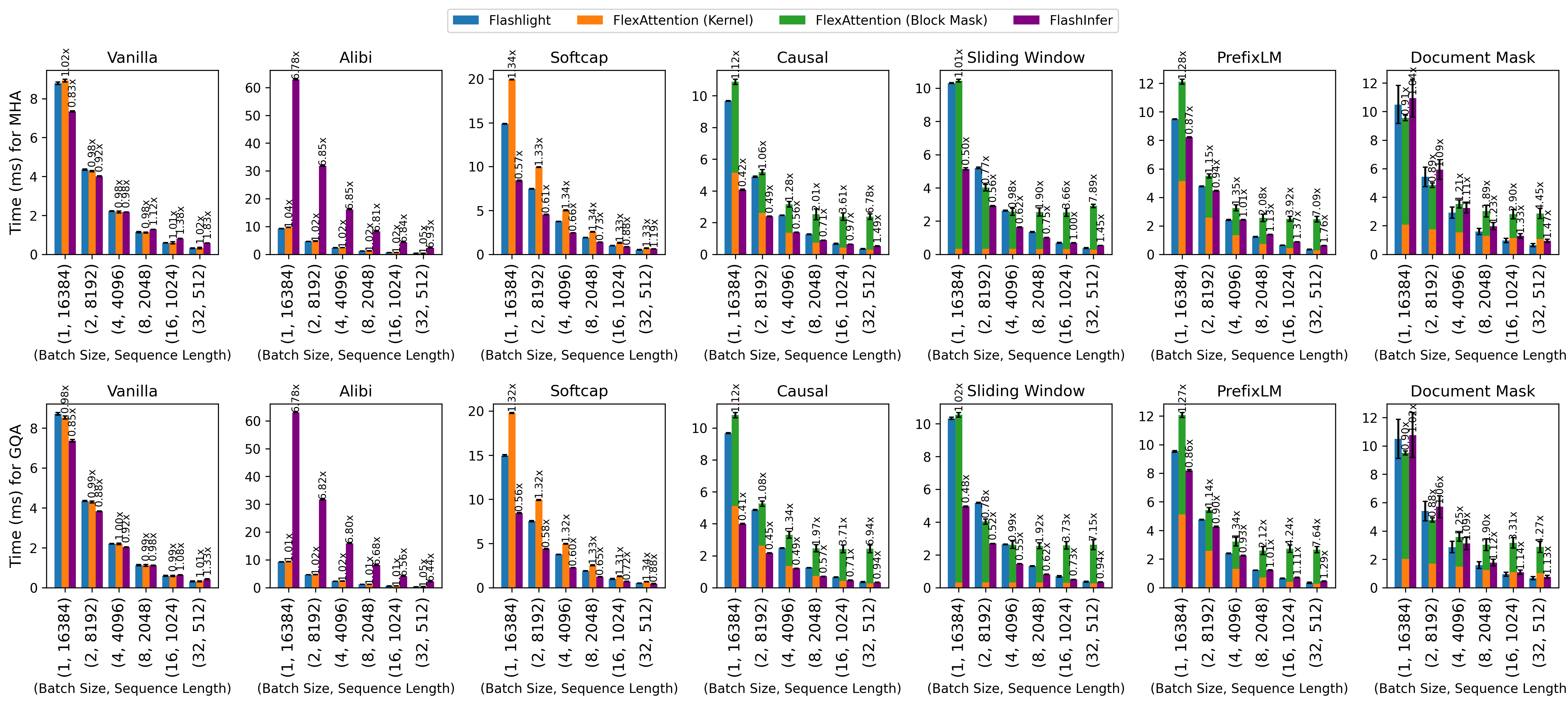}
  \caption{\revised{\flashlight{}, FlexAttention, and FlashInfer} on A100 for attention variants that are supported by FlexAttention template.}
  \label{fig:flex-able-a100}
\end{minipage}
\end{figure*}

\begin{figure*}
\centering
\includegraphics[width=0.98\linewidth]{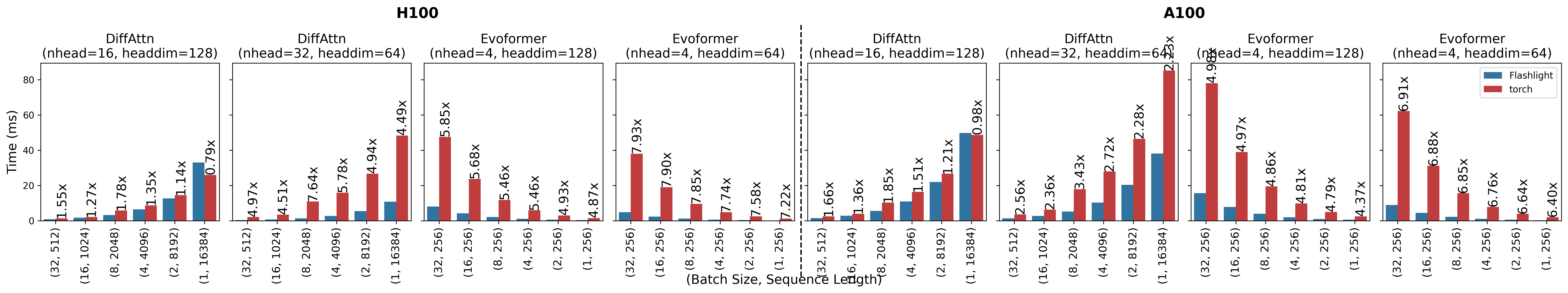}
\caption{Runtimes of \flashlight{} and torch.compile on H100/A100 for attention variants that are not supported by FlexAttention.}
\label{fig:flex-unable}
\end{figure*}


\subsection{Experimental Setup}
We run our experiments on two recent GPUs: an NVIDIA H100 80GB and NVIDIA A100 80GB.
We use Python 3.12, PyTorch 2.5.0, Triton 3.1.0, CUDA 12.9\revised{, FlashInfer 0.2.5}.
\revised{Except for end-to-end inference latency experiments (Section~\ref{subsec:inference}),}
we report average runtimes of 20 runs after 10 warm up runs.
To ensure that runs are within 1\% standard deviation, we cap the SM frequency of \revised{H100 to 1290 MHz and A100 to 1080 MHz}, which is the steady state frequency of the GPUs in our systems. 
%

{\bf Systems:}
We use these systems in our evaluation.
\emph{FlexAttention}~\cite{dong2024flexattentionprogrammingmodel} compiles attention code written in their template API in PyTorch to a fused Triton kernel.
\revised{\emph{FlashInfer}~\cite{ye2025flashinferefficientcustomizableattention} is a code-generation based attention engine in CUDA; we use it as a representative of optimized and tuned CUDA/C++ attention kernels.}
\emph{torch.compile}~\cite{Ansel2024pytorch2} compiles idiomatic attention code in PyTorch to multiple Triton kernels.
\emph{\flashlight{}} is our system described in the paper that compiles the same PyTorch code to a fused Triton kernel.

{\bf Benchmarks:}
We use these attention variants that are supported by FlexAttention: \emph{Vanilla}~\cite{vaswani2017attention}, \emph{ALiBi}~\cite{press2022trainshorttestlongalibi}, \emph{Softcap}~\cite{dong2024flexattentionprogrammingmodel}, \emph{Causal}~\cite{dong2024flexattentionprogrammingmodel}, \emph{Sliding Window}~\cite{beltagy2020longformerlongdocumenttransformer}, \emph{PrefixLM}~\cite{dong2024flexattentionprogrammingmodel}, and \emph{Document Mask}~\cite{dong2024flexattentionprogrammingmodel}.
Each of these variants are evaluated as Multi-Head Attention (MHA) and Grouped Query Attention (GQA).
For FlexAttention, \emph{Vanilla}, \emph{ALiBi}, and \emph{Softcap} use \texttt{score\_mod}, while the rest compute and use a \texttt{block\_mask}.
Like FlashAttention~\cite{dao2023flashattention2fasterattentionbetter}, we vary the sequence length from 512 to 16k and set the batch size so that their product (the number of tokens) is 16k. The head dimension is 64. For MHA, the number of heads is 16 for $Q$, $K$, and $V$; for GHA, the number of heads is 16 for $Q$ and 2 for $K$ and $V$. 
The sliding window size and the prefix length is 256. The number of documents is 12.

We also use attention variants that are not supported by FlexAttention:
differential attention~\cite{ye2024differentialtransformer} and row-wise gated self-attention in AlphaFold's Evoformer~\cite{alpafold}. 
We will call these DiffAttn and Evoformer respectively in the rest of this section. 
DiffAttn is shown in Listing~\ref{lst:diffattn}.
Evoformer uses an additional (sequence length) dimension and adds two bias matrices before $softmax$, one of which needs to be broadcasted along that dimension.
For DiffAttn, we use the same configuration for the MHA variants above except that we also evaluate 16 heads and 128 head dimension. 
For Evoformer, we vary the batch size from 1 to 32 and use 256 for the two sequence length dimensions; we use 4 heads and evaluate 64 and 128 head dimensions.

\revised{
{\bf Models:} 
To evaluate end-to-end inference, we use two models: AlphaFold2~\cite{alpafold} and LLaMa-3.2-1B~\cite{herd2024llama3}. AlphaFold2 is a scientific model used to predict the structure of proteins and it contains attention variants that are not supported by FlexAttention. 
We modify attention in LLaMa-3.2-1B, a large language model, to variants that are supported by FlexAttention. 
}


\subsection{FlexAttention-Supported Attention Variants}
\label{subsec:flex-variants-results}

Figures~\ref{fig:flex-able-h100} and~\ref{fig:flex-able-a100} show the runtimes for FlexAttention-supported attention variants in FlexAttention, FlashInfer, and \flashlight{} on H100 and A100 respectively. 

The runtime for FlexAttention is split into \emph{Block-Mask} creation and \emph{Kernel} execution times. The text on the bars show the speedup of \flashlight{} over FlexAttention. 
FlexAttention is marginally faster than \flashlight{} for Vanilla in some cases and for batch size 1 of ALiBi MHA on H100. In all the other cases, \flashlight{} is similar or much faster. 
For \texttt{score\_mod} variants, \flashlight{} may be up to $1.48\times$ faster because the fused Triton kernel is simpler than that of FlexAttention because it does not have compute or memory instructions needed for handling full, partial, or empty blocks in FlexAttention's templatized kernel. 
For \texttt{block\_mask} variants, FlexAttention's \emph{Kernel} execution is always faster than \flashlight{}'s execution because it skips redundant computation \revised{by using a sparse block mask}. However, FlexAttention's \emph{Block-Mask} execution time \revised{to construct the sparse block mask} is much slower. This can be amortized over multiple calls by using a cache, but this depends on the workload. 
\revised{In Section~\ref{subsec:inference}, we evaluate an end-to-end inference workload to quantify this.}

\revised{
For almost all batch sizes and sequence lengths,
FlashInfer is faster than both FlexAttention and \flashlight{} for all variants except \emph{ALiBi}. 
\flashlight{} does not optimize for block (mask) sparsity, 
while FlashInfer optimizes for block sparsity 
without needing to materialize sparse data structures
as it passes specialized parameters (e.g., \texttt{causal} for causal or \texttt{window\_left} for sliding window) directly into its API (e.g., \texttt{plan()}), which the CUDA kernel evaluates inline. 
In contrast, while FlexAttention employs an LRU cache (keyed on tensor shapes and mask definitions) to construct and reuse sparse block masks across forward passes, the attention kernel must still fetch and evaluate this block mask from memory to determine sparsity. 
On the other hand, Flashlight and FlexAttention achieve higher performance than FlashInfer for \emph{ALiBi} positional encodings. 
Due to FlexAttention's compilation model, PyTorch's \texttt{torch.compile} (via Triton) seamlessly fuses the \texttt{score\_mod} function into the kernel. For example, the ALiBi slope calculation in FlexAttention is evaluated at compile-time or folded into fast in-register math.
The FlashInfer implementation, however, either computes the bias element-wise with high overhead, or passes the pre-computed ALiBi slopes as a separate buffer to the pre-compiled FlashAttention backend, incurring a global memory read penalty per block that FlexAttention avoids.
}

We omit torch.compile in Figures~\ref{fig:flex-able-h100} and~\ref{fig:flex-able-a100} because they are much slower than FlexAttention \revised{and FlashInfer} in almost all cases \revised{(these results are included in Figures~\ref{fig:torch-compile-h100} and~\ref{fig:torch-compile-a100} in Appendix~\ref{sec:appendix-torch-compile})}. The only exceptions are: 
\revised{(1) \emph{ALiBi}, where FlashInfer is slower due to the overheads described above; and}
(2) all \texttt{block\_mask} variants for batch size 16 and 32, where FlexAttention is slower due to the overheads in creating the \texttt{block\_mask}.
\revised{In both cases, \flashlight{} is still faster than torch.compile.}

\subsection{Complex Attention Variants}

For attention variants that are not supported by FlexAttention,
Figure~\ref{fig:flex-unable}  show the runtimes of \flashlight{} and torch.compile on H100 and A100.
\flashlight{} is always faster than torch.compile. For DiffAttn, the speedup of \flashlight{} over torch.compile is higher on H100 than on A100.
For Evoformer, the speedups are $5\times$ or more on both H100 and A100. 

\subsection{End-to-End Inference Latency}
\label{subsec:inference}

\revised{To evaluate end-to-end inference for models that contain complex attention variants that are not supported by FlexAttention,}
we use the AlphaFold2~\cite{alpafold} model in the OpenFold~\cite{openfold} repo. The model has 48 Evoformer layers. We evaluate its inference latency with a sequence length of 256 (for both sequence length dimensions) and vary the batch size (the number of sequences) from $1, 2, 4, ..., 32$. 
Evoformer uses 8 heads and head dimension 32; Invariant Point Attention uses 12 heads and head dimension 16.
We evaluate PyTorch as-is without using the compiler. We add a \texttt{torch.compile} step to the row-wise and column-wise gated self attentions in Evoformer and then evaluate it without and with \flashlight{}. 
There is negligible difference in inference latency between PyTorch and torch.compile. 
\flashlight{} improves the inference latency by $6\%$ to $9\%$ on both H100 and A100 GPUs. 

\begin{figure}
\centering
\includegraphics[width=0.98\linewidth]{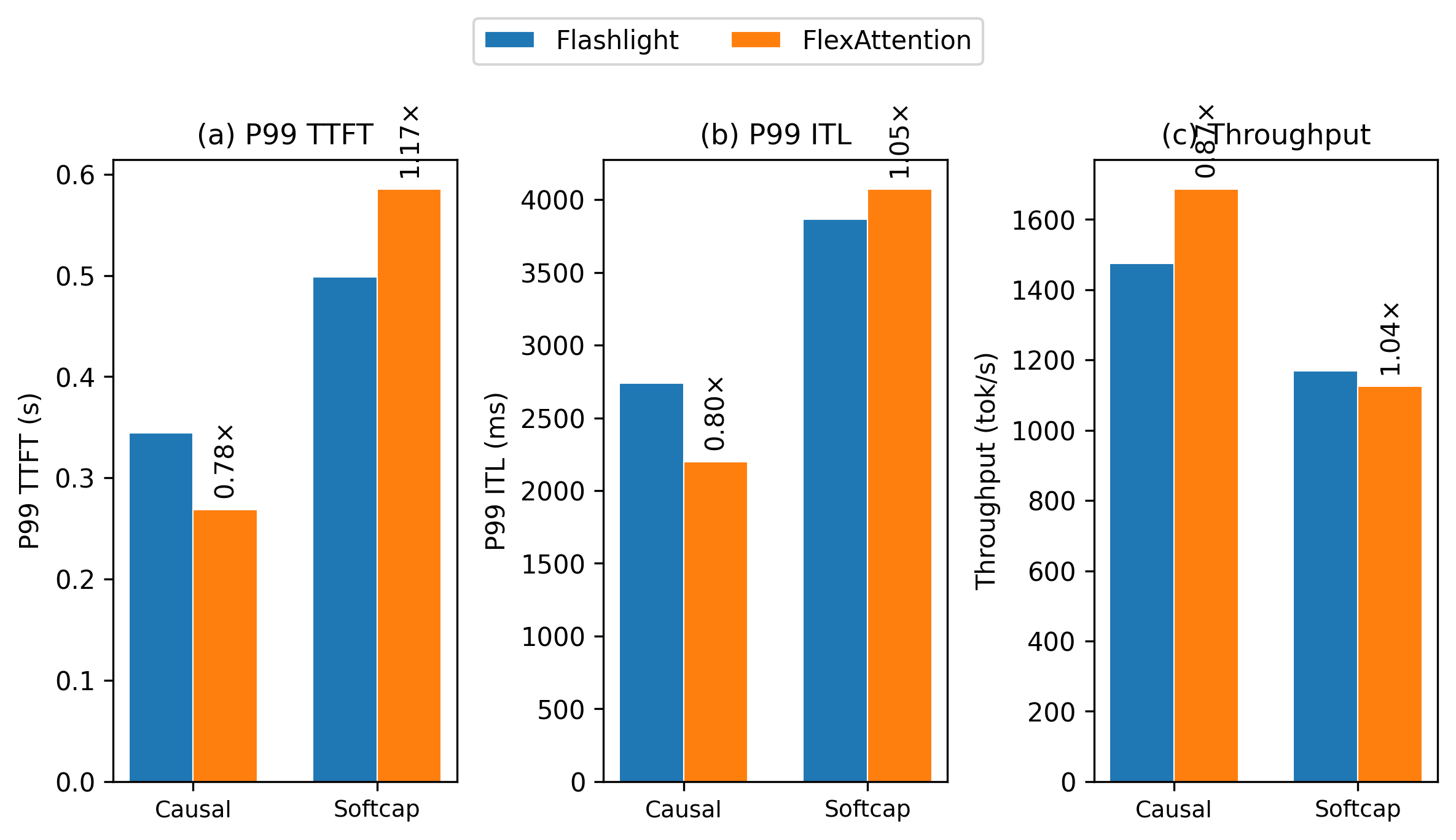}
\caption{\revised{Mooncake conversation trace in vLLM for LLaMa-3.2-1B inference with different attention variants using \flashlight{} and FlexAttention on H100.}}
\label{fig:mooncake-vllm}
\end{figure}

\revised{
To evaluate end-to-end inference for models that contain attention variants that are supported by FlexAttention, we use LLaMa-3.2-1B~\cite{herd2024llama3}, 
which uses \emph{Vanilla} attention. 
We modify it to use \emph{Causal} and \emph{Softcap} attention variants generating two new model variants. We use the Mooncake~\cite{Qin2025Mooncake} conversation trace to evaluate these models in the vLLM system~\cite{kwon2023vLLM} on the H100 GPU. We run only the first 200 requests in the trace (after a warmup run of the same requests). Figure~\ref{fig:mooncake-vllm} shows the time to first token latency (TTFT), inter-token latency (ITL), and token throughput. For TTFT and ITL, lower is better; for token throughput, higher is better. 
Note that PyTorch and default torch.compile run out-of-memory for these model variants 
because the attention variants are not fused (which leads to the materialization of the intermediate attention tensor).
\flashlight{} is more performant than FlexAttention for \emph{Softcap} as its kernel execution time is faster than that of FlexAttention, as shown in Figure~\ref{fig:flex-able-h100}. 
On the other hand, for \texttt{block\_mask} variants like \emph{Causal}, FlexAttention's kernel execution time is faster than that of \flashlight{} and its sparse block mask creation time is amortized over multiple calls to the kernel (for the same tensor shapes), so FlexAttention is more performant. 
Extending \flashlight{} to better optimize for such structured block sparsity is left for future work.
}

\section{Related Work}\label{sec:related_work}


{\bf Efficient Attention Kernels.}
\textbf{FlashAttention}~\cite{dao2022flashattentionfastmemoryefficientexact,dao2023flashattention2fasterattentionbetter,shah2024flashattention3fastaccurateattention} introduced tiling and kernel fusion to reduce memory traffic while keeping exact results. \textbf{FlexAttention}~\cite{dong2024flexattentionprogrammingmodel} extended this idea with static templates for specialized attention patterns, and \textbf{FlashInfer}~\cite{ye2025flashinferefficientcustomizableattention} applied similar techniques to speed up LLM inference. Unlike these systems, \textbf{Flashlight} is built directly into the PyTorch compiler, automatically generating optimized kernels during \texttt{torch.compile}.

{\bf Flexible or Specialized Attention Models.}
Recent work on attention variants, including \textbf{Longformer}~\cite{beltagy2020longformerlongdocumenttransformer}, 
\textbf{ALiBi}~\cite{press2022trainshorttestlongalibi}, and \textbf{DiffTransformer}~\cite{ye2024differentialtransformer}, demonstrated how to improve efficiency and robustness through sparsity, positional biasing, or continuous updates. These motivate the need for a general compiler framework that could automatically optimize new attention types without hand-written kernels.

{\bf Compiler Infrastructure for Deep Learning.}
\textbf{Triton}~\cite{Tillet2019Triton} provides the GPU DSL used by TorchInductor. 
\revised{\textbf{PolyBlocks}~\cite{polyblocks}, a concurrent work, is a MLIR-based compiler that fuses attention using affine access analysis.}
Other compiler frameworks, such as \textbf{TVM}~\cite{chen2018tvm}, which relies on autotuning, and \textbf{Mirage}~\cite{wu2025mirage}, which uses program synthesis to generate new kernels, have similar goals, but they are outside the PyTorch ecosystem. DSLs like \textbf{ThunderKittens}~\cite{spector2024thunderkittens} explore kernel fusion and scheduling for simpler and faster GPU programming. 


\section{Conclusion and Future Work}
The success of attention in LLMs has inspired researchers to design many other attention models, but these are not supported well in existing frameworks. \flashlight{} is a compiler-native framework within the PyTorch ecosystem that automatically generates fused, FlashAttention-style kernels for arbitrary attention programs, without relying on static templates or predefined kernel specializations. It supports all attention variants expressible in the FlexAttention model but also handles more general, data-dependent attention formulations beyond the capabilities of FlexAttention. Our results show that \flashlight{} produces
kernels with competitive or superior performance to expert-tuned implementations, while offering the flexibility
of native PyTorch code, enabling developers to rapidly explore new attention models
without sacrificing performance.
\revised{While we evaluated \flashlight{} only for attention variants in AI inference, the extensions and passes in \flashlight{} are general-purpose and can enable fusion in other use cases. 
\flashlight{} could be extended to support training workloads and to optimize for block sparsity in attention.}
\section*{Acknowledgements}


The UT Austin team was supported in part by NSF grant 2505085.
This project is partially supported by the Natural Sciences and Engineering Research Council of Canada (NSERC) [funding reference number 587440-2024]. 
This research was supported through cyber-infrastructure research resources and services provided by the Partnership for an Advanced Computing Environment (PACE) at the Georgia Institute of Technology.
\newpage

\bibliographystyle{mlsys2026}
\bibliography{structural/references}

\clearpage
\appendix

\section{Algebraic Transformation of Reductions}
\label{sec:alg-reduction-appendix}

The online softmax algorithm~\cite{milakov2018onlinesoftmax} (Alg.~\ref{alg:online_softmax}) is key to implementing FlashAttention-like fused kernel.
Machine learning developers have to explicitly replace the standard stable softmax implementation with online softmax. However, modern compilers for machine learning frameworks, including the PyTorch compiler, do not currently detect a pattern as such and therefore do not generate an online implementation automatically.We show that the conversion of the stable softmax algorithm to the online softmax algorithm can be generalized using the standard algebraic notion of a {\em homomorphism}. Informally, a homomorphism is a structure-preserving map between two algebraic structures of the same type such as two groups.

\begin{definition}
\label{def:homomorphism}
Let $A$ be a set with a binary operation $\oplus$, and let $B$ be a set with a binary operation $\otimes$. A function $f:A{\rightarrow}B$ 
is said to be a homomorphism if for all $a_1,a_2 \in A$, $f(a_1\oplus a_2) = f(a_1)\otimes f(a_2)$. 
\end{definition}


In the context of softmax, $A = B = \mathbb{R}$ (the set of real numbers), and $\oplus$ and  $\otimes$ are addition ($+$)  and multiplication ($\times$) of real numbers. The function $f(x) = e^x$ is a homomorphism because $f^{a+b} = f^a\times f^b$. To generalize the online softmax construction, we need the set $A$ with operations $\oplus$ and $\otimes$ to satisfy the axioms of a {\em ring}:

\begin{itemize}
\item $\oplus$ is associative, there is an element $0\in A$ such that $a\oplus 0 = 0\oplus a = a$, and every element $a$ has an additive inverse, denoted by $\ominus a$, such that $a\oplus (\ominus a) = (\ominus a)\oplus a = 0$.

\item $\otimes$ is associative, and there is an element $1\in A$ such that $a\otimes 1 = 1\otimes a = a$.

\item $\otimes$ distributes over $\oplus$; that is, $(a\oplus b) \otimes c = (a\otimes c) \oplus (b \otimes c)$.     
\end{itemize}

It can be shown that these assumptions imply $a\otimes 0 = 0 \otimes a = 0$. The standard definition of a ring requires $\oplus$ to be commutative but we do not use this property in the development below. In the context of rings, a homomorphism $f$ must also satisfy $f(0) = 1$.

In the stable softmax algorithm (Alg.~\ref{alg:stable_softmax}), let us denote the sequence of $m$ values produced by the first loop by $m[1..N]$ and let $m[0]=0$ by definition. The sequence of $d$ values produced by the second loop, which we denote by $ds$, is expressed abstractly by the following recurrence in which the elements of $ds$ are members of a ring $A$ and $E{:}A{\rightarrow}A$ is a homomorphism.  

{\scriptsize
\begin{mdframed}
\begin{flalign*}
ds[0] &= 0 &\\
ds[j] &= ds[j{-}1] \oplus (E(x[j]\oplus(\ominus m[N]))) \ ~|N\geq j \geq 1 &\\
&= ds[j{-}1] \oplus (E(x[j])\otimes E(\ominus m[N])) \ ~|N\geq j \geq 1 &
\end{flalign*}
\end{mdframed}
}

It is easy to show by induction that $ds$ can be expressed in closed-form by the following expression in which $\bigoplus$ stands for the application of the associative operation $\oplus$ to a set of elements of $A$.

{\scriptsize
\begin{mdframed}
\allowdisplaybreaks
\begin{flalign*}
ds[j] &= \bigoplus_{i{=}1}^j (E(x[i]\oplus(\ominus m[N]))) \ ~|N\geq j \geq 1 & \\
&= \bigoplus_{i{=}1}^j (E(x[i])\otimes E(\ominus m[N])) \ ~|N\geq j \geq 1 & \\
&\ \ \ \ \text{($E$ is a homomorphism)} &\\
\begin{split}
= \bigg(\bigoplus_{i{=}1}^j E(x[i])\bigg)  \otimes E(\ominus m[N])  \\
\ \ \ \ \text{(from distributivity of $\otimes$ over $\oplus$)} 
\end{split}
\end{flalign*}
\end{mdframed}
}

The online softmax algorithm (Alg.~\ref{alg:online_softmax}) computes a different sequence, denoted by $do$, that can be expressed abstractly as shown below. 

{\scriptsize
\begin{mdframed}
\vspace*{-0.1in}
\begin{flalign}
& do[0] = 0  \\
\begin{split}
do[j] = \bigg(do[j{-}1]\otimes E(m[j{-}1] \oplus (\ominus m[j]))\bigg) \oplus  \\
(E(x[j]\oplus (\ominus m[j]))) ~|N\geq j \geq 1
\label{doDef}
\end{split}&\\
\begin{split}
\ \ \ \ \ \ \ = \bigg(do[j{-}1]\otimes E(m[j{-}1]) \otimes E(\ominus m[j])\bigg) \oplus  \\
(E(x[j]\oplus (\ominus m[j]))) ~|N\geq j \geq 1 \\
({\rm from\ distributivity\ of\ \otimes\ over\ \oplus})
\label{doDef}
\end{split}&
\end{flalign}
\end{mdframed}
}

The $ds$ and $do$ sequences will be different in general, but we show that $do$ can be expressed in closed-form by the following expression.

{\scriptsize
\begin{mdframed}
\vspace*{-0.1in}
\begin{flalign}
do[j]  & = \bigg(\bigoplus_{i{=}1}^j E(x[i])\bigg) \otimes E(\ominus m[j])  \ ~|N \geq j \geq 1 & \label{doClosed}
\end{flalign}
\end{mdframed}
}
from which it follows that $ds[N] = do[N]$. 
The proof of correctness of (\ref{doClosed}) is by induction on $j$. 
\begin{itemize}[leftmargin=-0.1em]
\item $j{=}1$: From (\ref{doDef}),

\allowdisplaybreaks
{\scriptsize
\begin{mdframed}
\begin{flalign}
do[1] &= (do[0]\otimes E(m[0])\otimes E(\ominus m[1]))\oplus (E(x[1])\otimes E(\ominus m[1]) ) \\
&= E(x[1])\otimes E(\ominus m[1]) \text{\ \ (since $do[0] = 0)$} 
\end{flalign}
\end{mdframed}
}

This is the value obtained from (\ref{doClosed}) for $j=1$.

\item $j{>}1$: Assume inductively that 

{\scriptsize
\[
do[j{-}1] = \bigg( \bigoplus_{i{=}1}^{j{-}1} E(x[i]) \bigg) \otimes E(\ominus m[j{-}1]) 
\]}

From (\ref{doDef}), 

{\scriptsize
\begin{mdframed}
\begin{align*}
do[j] &= \bigg(do[j{-}1]\otimes E(m[j{-}1])\otimes E(\ominus m[j])\bigg) \oplus \\& \bigg(E(x[j])\otimes E(\ominus m[j])\bigg) \\
&= \bigg(\bigg( \bigoplus_{i{=}1}^{j{-}1} E(x[i]) \bigg) \otimes \underbrace{E(\ominus m[j{-}1]) \otimes E(m[j{-}1])}_{\text{=1 because $E$ is homomorphism}}\otimes  \\& E(\ominus m[j])\bigg) \oplus \bigg(E(x[j])\otimes E(\ominus m[j])\bigg) \\
&= \bigg(\bigg( \bigoplus_{i{=}1}^{j{-}1} E(x[i]) \bigg) \otimes E(\ominus m[j])\bigg) \oplus  \\& \bigg(E(x[j])\otimes E(\ominus m[j])\bigg) \\
&= \bigg(\bigg( \bigoplus_{i{=}1}^{j{-}1} E(x[i]) \bigg) \oplus E(x[j]\bigg)\otimes E(\ominus m[j])  \\& \text{\ \ (from distributivity of $\otimes$ over $\oplus$)} \\
&= \bigg( \bigoplus_{i{=}1}^{j} E(x[i]) \bigg) \otimes E(\ominus m[j]) \text{\ \ (as required by (\ref{doClosed}))}
\end{align*}
\end{mdframed}
}
\end{itemize}


\section{\texttt{torch.compile} Results}
\label{sec:appendix-torch-compile}

\begin{figure*}
\centering
\includegraphics[width=0.98\linewidth]{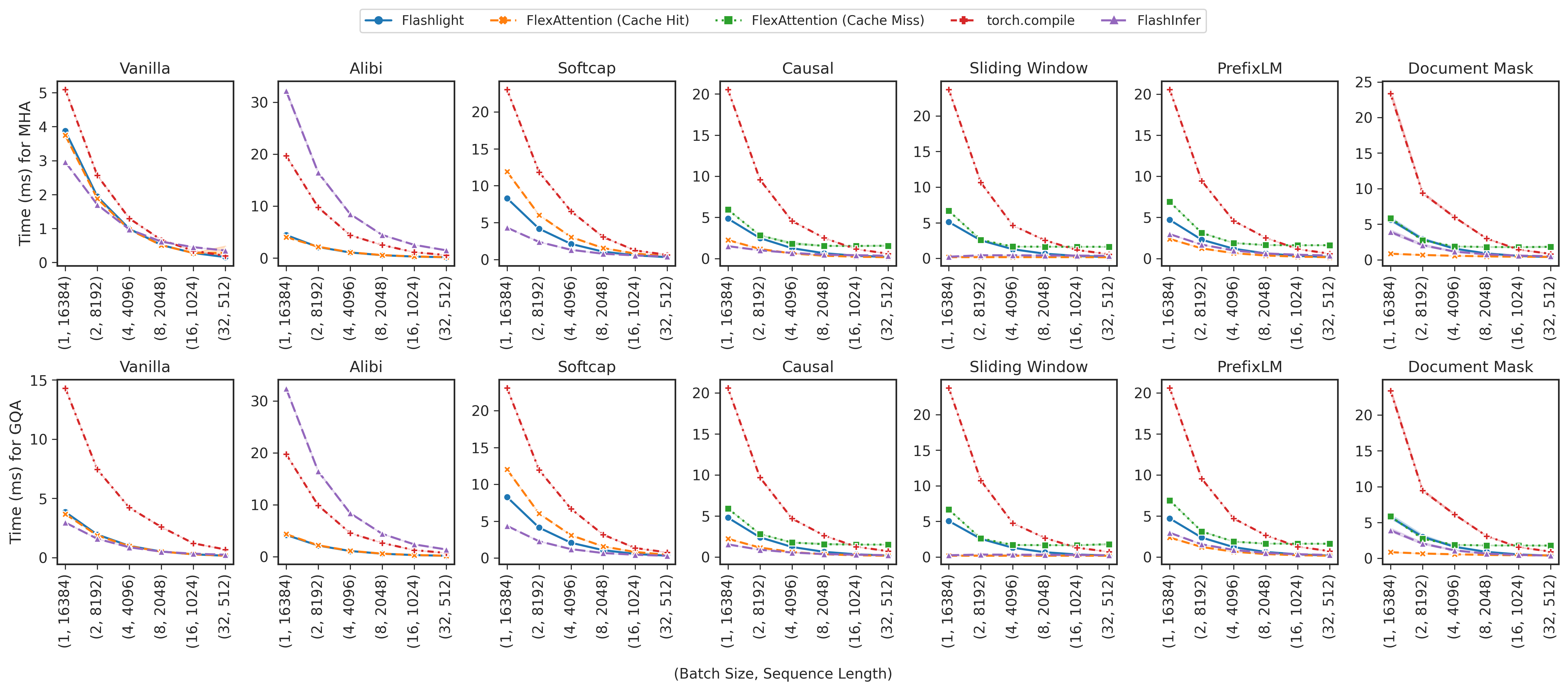}
\caption{Execution times on H100 for attention variants that are not supported by FlexAttention.}
\label{fig:torch-compile-h100}
\end{figure*}

\begin{figure*}
\centering
\includegraphics[width=0.98\linewidth]{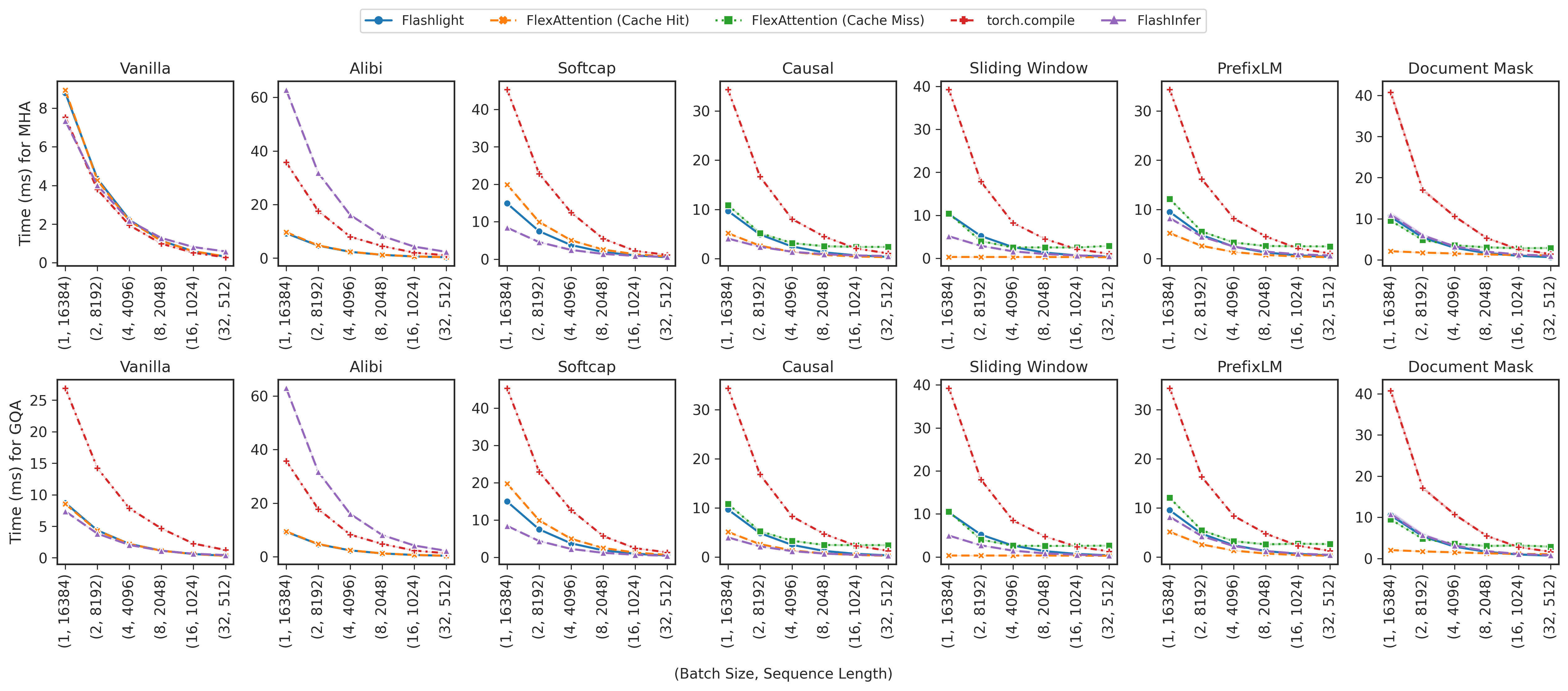}
\caption{Execution times on A100 for attention variants that are not supported by FlexAttention.}
\label{fig:torch-compile-a100}
\end{figure*}

\revised{
Figures~\ref{fig:torch-compile-h100} and~\ref{fig:torch-compile-a100} shows the execution times for torch.compile along with FlashInfer, FlexAttention, and \flashlight{} on H100 and A100 GPU respectively. In almost all cases, torch.compile is slower than the others. The only exceptions are:
(1) \emph{ALiBi}, where FlashInfer is slower due to the overheads described in Sectin~\ref{subsec:flex-variants-results}; and
(2) all \texttt{block\_mask} variants for batch size 16 and 32, where FlexAttention is slower due to the overheads in creating the \texttt{block\_mask}.
In both cases, \flashlight{} is still faster than torch.compile.
}
\section{Artifact Appendix}

\subsection{Abstract}

The artifact for \textit{\flashlight: PyTorch Compiler Extensions to Accelerate Attention Variants} provides the source code, benchmark scripts, and environment definitions needed to reproduce the experimental results presented in the paper. \flashlight is a compiler-native framework within the PyTorch ecosystem that automatically generates fused, FlashAttention-style kernels for a wide range of attention programs, including data-dependent variants that go beyond what FlexAttention can express. The implementation is delivered as a monkey patch on top of stock PyTorch~2.5.0 so that it can be dynamically imported for side-by-side benchmarking against the original framework. The artifact packages the evaluation pipelines for FlexAttention-supported variants and complex variants such as Differential Attention and Evoformer, together with an Apptainer container definition and a local \texttt{uv}-based runner. Evaluators can run the provided workflows to collect per-kernel forward runtimes and derived TFLOPs and automatically regenerate the bar charts corresponding to the paper's figures on NVIDIA A100 or H100 GPUs.

\subsection{Artifact check-list (meta-information)}

{\small
\begin{itemize}
  \item {\bf Algorithm: } \flashlight fusion for FlexAttention-compatible variants (Vanilla, ALiBi, Softcap, Causal, Sliding Window, Prefix LM, Document Mask) plus Differential Attention (DiffAttn) and Evoformer-style attention (see Section~\ref{sec:eval}).
  \item {\bf Program: } PyTorch~2.5.0 with \flashlight monkey-patch extensions; benchmark drivers under \texttt{benchmarks/} and \texttt{attention\_variants/}; Apptainer-based and local \texttt{uv}-based runners.
  \item {\bf Compilation: } PyTorch compilation stack (TorchInductor) with Triton~3.1.0 for GPU JIT compilation of fused attention kernels; no additional host C/C\texttt{++} or \texttt{nvcc} build steps beyond those used by the PyTorch distribution.
  \item {\bf Binary: } Generated CUDA kernels via PyTorch/Triton; Apptainer image \texttt{flashlight.sif}.
  \item {\bf Data set: } Synthetic, randomly generated tensors for kernel benchmarks; conversation trace from \url{https://github.com/kvcache-ai/Mooncake/blob/3cca71d/FAST25-release/traces/} for end-to-end benchmark.
  \item {\bf Run-time environment: } Python~3.12; PyTorch~2.5.0; Triton~3.1.0; FlashInfer 0.2.5; vLLM 0.6.6; CUDA~12.9 (matching the evaluation setup in Section~\ref{sec:eval}); Apptainer~1.4.1 on TACC Lonestar6 tested but not required, as long as the host driver supports the chosen PyTorch CUDA wheels.
  \item {\bf Hardware: } 1$\times$ NVIDIA A100~80GB or H100~80GB; other recent NVIDIA GPUs may exhibit different absolute runtimes.
  \item {\bf Execution: } End-to-end execution via \texttt{make -C benchmarks} invoked either inside the Apptainer image or from the local \texttt{uv} environment.
  \item {\bf Metrics: } Per-kernel forward runtime in milliseconds and derived TFLOPs for attention kernels (as reported by the benchmark scripts); for the AlphaFold case study, end-to-end inference latency.
  \item {\bf Output: }
    \begin{itemize}
        \item CSV results in \texttt{benchmarks/results/*.csv}
        \item PNG plots in \texttt{benchmarks/results/*.png}.
    \end{itemize}
  \item {\bf How much disk space required (approximately)?: } $\sim$5 GB for Python virtual environment or $\sim$7 GB for Apptainer SIF image.
  \item {\bf How much time is needed to complete experiments (approximately)?: } $\sim$15 minutes on a single NVIDIA A100-class GPU for a full end-to-end run of \texttt{./scripts/run\_mlsys26\_ae\_local.sh} with default configurations (timings on other recent NVIDIA GPUs may vary slightly, but relative trends remain unchanged).
  \item {\bf Publicly available?: } Yes. Rrimary code repository is available at \url{https://github.com/bozhiyou/flashlight/tree/mlsys26-ae}.
  \item {\bf Code licenses?: } BSD 3-Clause License (see \texttt{LICENSE} in the repository).
  \item {\bf Data licenses (if publicly available)?: } Not applicable (data used in kernel benchmarks are synthetic and generated on the fly; the end-to-end benchmark uses the Mooncake conversation trace).
  \item {\bf Archived (provide DOI)?: } \url{https://doi.org/10.5281/zenodo.18990626}
\end{itemize}
}

\subsection{Description}

\subsubsection{How delivered}

The artifact is delivered as a public GitHub repository (branch/tag \texttt{mlsys26-ae}). It includes the \flashlight compiler extensions (\texttt{monkeypatch/}) implemented as a monkey patch over PyTorch~2.5.0, benchmark implementations (\texttt{benchmarks/}), variant definitions (\texttt{attention\_variants/}), and environment setup files (\texttt{apptainer/} and \texttt{scripts/}). An archived snapshot, suitable for long-term availability and badging, is provided on Zenodo under DOI \url{https://doi.org/10.5281/zenodo.18990626}.

\subsubsection{Hardware dependencies}

The benchmarks are designed for a single \textbf{NVIDIA A100 80GB} or \textbf{NVIDIA H100 80GB} GPU, matching the systems used in the paper's evaluation (Section~\ref{sec:eval}). For consistent reproduction of the reported speedups, it is strongly recommended to lock the GPU SM frequency (\texttt{sudo nvidia-smi ---lock-gpu-clocks=<frequency>}) as described in the experimental setup of Section~\ref{sec:eval}; see the frequency-capping note in Section~\ref{sec:ae-notes}.

\subsubsection{Software dependencies}

Reviewers can choose between two deployment methods:
\begin{itemize}
    \item \textbf{Apptainer (recommended on shared clusters such as TACC Lonestar6)}: \texttt{Apptainer}~1.4.1 and a host NVIDIA driver compatible with CUDA~12.1 (the configuration used in our tests); any environment that can run the built image with GPU support (\texttt{apptainer run --nv}) is acceptable.
    \item \textbf{Local Environment (via \texttt{uv})}: Requires Python >3.10 and <3.13 on the host, \texttt{uv}, \texttt{git}, \texttt{make}, an NVIDIA driver compatible with the chosen \texttt{torch==2.5.0} CUDA wheel (e.g., CU121), and internet access to install dependencies (including a pinned commit of \texttt{attention-gym}) as automated by \texttt{scripts/run\_mlsys26\_ae\_local.sh}.
\end{itemize}

\subsubsection{Data sets}

The kernel benchmarks synthesize random tensors based on the configured batch sizes, sequence lengths, and hidden dimensions.
The end-to-end benchmark uses the Mooncake conversation trace, available at \url{https://github.com/kvcache-ai/Mooncake/blob/3cca71d/FAST25-release/traces/}.

\subsection{Installation}

We provide two seamless workflows to initialize the environment and run the evaluation:

\textbf{Option A: Apptainer (Containerized)} \\
From the repository root, follow \texttt{apptainer/README.md}. On TACC Lonestar6 this typically looks like:
\begin{verbatim}
module load tacc-apptainer/1.4.1
make -C apptainer
\end{verbatim}
which builds the \texttt{flashlight.sif} image and then runs the artifact evaluation via the container's default runscript (invoking \texttt{make -C benchmarks all} inside the image).

\textbf{Option B: Local Environment} \\
We provide a setup script that creates an isolated \texttt{uv} virtual environment, installs PyTorch~2.5.0 and all dependencies (including a pinned version of \texttt{attention-gym}), and runs the benchmarks and plotting:
\begin{verbatim}
./scripts/run_mlsys26_ae_local.sh
\end{verbatim}
This is the recommended entry point for reviewers who do \emph{not} have access to TACC or Apptainer: running the single command above on a CUDA-capable machine is sufficient to set up the environment and reproduce the figures locally.
The script mirrors the container workflow: it sets up the environment, ensures \texttt{PYTHONPATH=.} so that the \flashlight monkey patches are importable, exports \texttt{FL\_GPU\_CLOCK\_FREQ\_MHZ} (default 1290 MHz), and finally calls \texttt{make -C benchmarks all} from the repository root.

\subsection{Experiment workflow}

The automated scripts (the \texttt{Makefile} in \texttt{apptainer/}, or \texttt{scripts/run\_mlsys26\_ae\_local.sh}) wrap the central benchmarking Makefile located at \texttt{benchmarks/Makefile}. The workflow consists of two phases:

\textbf{1. Data Generation (\texttt{make data}):}
Executes \texttt{run\_fig2\_fig3\_flex\_variants.py} to collect results for \flashlight and FlexAttention (cache hit and cache miss) and then runs \texttt{run\_fig4\_diff\_attn.py} and \texttt{run\_fig4\_evoformer.py} for DiffAttn and Evoformer. By default this produces (in \texttt{benchmarks/results}):
\begin{itemize}
  \item \texttt{all.csv} (\flashlight),
  \item \texttt{all\_flex.csv} (FlexAttention, cache hit),
  \item \texttt{all\_flexnocache.csv} (FlexAttention, cache miss),
  \item \texttt{all\_flashinfer.csv} (FlashInfer),
  \item \texttt{all\_torchcompile.csv} (torch.compile),
  \item \texttt{diff\_attn.csv} and \texttt{evo\_attn.csv} (DiffAttn and Evoformer),
  \item \texttt{vllm\_e2e\_online\_summary.csv} and \texttt{vllm\_e2e\_online\_per\_request.csv} (vLLM end-to-end inference),
\end{itemize}
Figure~\ref{fig:flex-unable} plots results on both A100 and H100 GPUs, but the artifact is run only on only one of them, so static reference CSVs under \texttt{benchmarks/results/reference/} are committed for completing Figure~\ref{fig:flex-unable}.

\textbf{2. Figure Plotting (\texttt{make figures}):}
Parses the generated CSV files using \texttt{plot\_fig2\_fig3.py} and \texttt{plot\_fig4.py} to output \texttt{fig2\_fig3.png} and \texttt{fig4.png} in \texttt{benchmarks/results/}. The bar charts in Figures~\ref{fig:flex-able-h100} or~\ref{fig:flex-able-a100} are produced by \texttt{plot\_fig2\_fig3.py} from these CSVs, while Figure~\ref{fig:flex-unable} is produced by \texttt{plot\_fig4.py}. For Figures~\ref{fig:flex-able-h100} and~\ref{fig:flex-able-a100}, the same pipeline is used on both A100 and H100; which figure is instantiated in the paper depends on the GPU on which the benchmarks are run. For Figure~\ref{fig:flex-unable}, if data for a specific GPU architecture is missing in the newly generated CSVs, the plotting script gracefully falls back to using the static reference CSVs provided in \texttt{benchmarks/results/reference/}.
Figure~\ref{fig:mooncake-vllm} is produced by \texttt{plot\_fig5.py}.

\textbf{Quick Sanity Check (Optional):}
To verify functionality without waiting 15 minutes, reviewers can manually invoke a smaller run from the repository root:
\begin{verbatim}
PYTHONPATH=. python \
  benchmarks/run_fig4_diff_attn.py \
  --batch_size 1 --seqlen 512
\end{verbatim}

\subsection{Evaluation and expected result}

\textbf{Functional:} The artifact passes the functional badge if the \texttt{Makefile} completes successfully and produces the output CSVs and PNG figures.

\textbf{Reproducible:} The artifact supports the "Results Reproduced" badge by yielding relative speedup trends that align with the paper:
\begin{itemize}
    \item \textbf{Figures~\ref{fig:flex-able-h100} and~\ref{fig:flex-able-a100}}: \flashlight is competitive with or faster than FlexAttention for \texttt{score\_mod} variants. For \texttt{block\_mask} variants, while the FlexAttention kernel alone may be faster, \flashlight outperforms the end-to-end FlexAttention pipeline which incurs block-mask creation overhead.
    \item \textbf{Figure~\ref{fig:flex-unable}}: \flashlight is consistently faster than standard \texttt{torch}. For Evoformer, it demonstrates speedups of $\sim$5$\times$ or more.
    \item \textbf{Figure~\ref{fig:mooncake-vllm}}: \flashlight is more performant than FlexAttention for \emph{Softcap}, while FlexAttention is more performant for \texttt{block\_mask} variants like \emph{Causal},.
  \end{itemize}
Due to inherent hardware variance, absolute runtime numbers and minor speedup fluctuations (typically within 1-5\%) are expected when comparing across different instances of the same GPU.

\subsection{Experiment customization}

Reviewers can customize the benchmark scope to explore specific claims:
\begin{itemize}
    \item \textbf{Filter Variants:} Pass the \texttt{--filter} flag to \texttt{run\_fig2\_fig3\_flex\_variants.py} to isolate specific attention variants (e.g., \texttt{--filter causal}).
    \item \textbf{Adjust Shapes:} Use \texttt{--batch\_size} and \texttt{--seqlen} to decrease the problem size for faster testing on memory-constrained GPUs.
    \item \textbf{Disable Mask Caching:} FlexAttention's mask cache overhead can be measured by running with \texttt{--flex --no-mask-cache}.
\end{itemize}

\subsection{Notes}
\label{sec:ae-notes}

\textbf{Why does frequency capping matter?} 
To ensure stable run-to-run measurements, the Evaluation section (Section~\ref{sec:eval}) locks the GPU Streaming Multiprocessor (SM) clock frequency (e.g., \texttt{sudo nvidia-smi -lgc <frequency>}). This configuration is strongly recommended whenever you have \texttt{sudo} access and is what we use for the paper's headline results. On shared clusters like TACC where \texttt{sudo} is unavailable, our scripts use a fallback heuristic: an environment variable \texttt{FL\_GPU\_CLOCK\_FREQ\_MHZ=1290} instructs the \texttt{warmup\_max} routine in \texttt{benchmarks/\_utils.py} to ``warm up'' the GPU until it naturally approaches the target frequency before recording times. This warmup-based approximation improves consistency relative to unconstrained runs but is inherently best-effort and may still introduce small discrepancies in absolute runtimes and speedups compared to fully frequency-capped runs reported in the paper.

\textbf{Compilation Overhead:} 
The first execution of \texttt{torch.compile} or \flashlight incurs substantial JIT compilation time (up to several minutes) as Triton kernels are generated and autotuned. Subsequent runs within the same script execution are fast.

\textbf{Inspecting fused kernels and TorchInductor cache:}
The fused Triton kernels and associated compiler artifacts generated by PyTorch and \flashlight are stored under the TorchInductor cache directory, controlled by the \texttt{TORCHINDUCTOR\_CACHE\_DIR} environment variable. By default, when this variable is unset, the cache is placed under a directory of the form \texttt{/tmp/torchinductor\_\$\{USER\}} on Linux systems. Reviewers interested in inspecting the generated kernels can browse this directory after running the benchmarks. If PyTorch compilation or \flashlight's compiler path encounters issues (for example, due to a stale or partially written cache), removing the contents of the TorchInductor cache directory and re-running the experiment is a recommended troubleshooting step.


\end{document}